\documentclass{article}

\usepackage[dvipsnames]{xcolor}
\usepackage{microtype}
\usepackage{graphicx}
\usepackage{subfigure}
\usepackage{booktabs} 
\usepackage{tabularx} 
\usepackage{arydshln} 
\usepackage{multirow} 
\usepackage{wrapfig} 
\usepackage{capt-of}
\usepackage{comment}

\usepackage{hyperref}

\usepackage[accepted]{icml2024}

\usepackage{amsmath}
\usepackage{amssymb}
\usepackage{mathtools}
\usepackage{amsthm}

\usepackage[capitalize,noabbrev]{cleveref}

\theoremstyle{plain}

\theoremstyle{definition}

\theoremstyle{remark}

\newcommand{\ourmethodlong}{\texttt{Sequential Autoencoder for Neural Embeddings}\,}
\newcommand{\ourmethod}{\texttt{SANE}\,}

\icmltitlerunning{Towards Scalable and Versatile Weight Space Learning}

\begin{document}

\twocolumn[
\icmltitle{Towards Scalable and Versatile Weight Space Learning}



\icmlsetsymbol{equal}{*}

\begin{icmlauthorlist}
\icmlauthor{Konstantin Schürholt}{unisg,icsi}
\icmlauthor{Michael W. Mahoney}{icsi,lbnl,ucb}
\icmlauthor{Damian Borth}{unisg}
\end{icmlauthorlist}

\icmlaffiliation{unisg}{AIML Lab, University of St.Gallen, St. Gallen, Switzerland}
\icmlaffiliation{icsi}{International Computer Science Institute, Berkeley, CA, USA}
\icmlaffiliation{ucb}{Department of Statistics, University of California at Berkeley, CA, USA}
\icmlaffiliation{lbnl}{Lawrence Berkeley National Laboratory, Berkeley, CA, USA}

\icmlcorrespondingauthor{Konstantin Schürholt}{konstantin.schuerholt@unisg.ch}

\icmlkeywords{Machine Learning, Weight Space, Representation Learning, ICML}

\vskip 0.3in
]



\printAffiliationsAndNotice{}  

\linepenalty=5000

\begin{abstract}
Learning representations of well-trained neural network models holds the promise to provide an understanding of the inner workings of those models.
However, previous work has either faced limitations when processing larger networks or was task-specific to either discriminative or generative tasks.
This paper introduces the \ourmethod approach to weight-space learning. 
\ourmethod overcomes previous limitations by learning task-agnostic representations of neural networks that are scalable to larger models of varying architectures and that show capabilities beyond a single task.
Our method extends the idea of \textit{hyper-representations} towards sequential processing of subsets of neural network weights, 
thus allowing one to embed larger neural networks as a set of tokens into the learned representation space. 
\ourmethod reveals global model information from layer-wise embeddings, and it can sequentially generate unseen neural network models, which was unattainable with previous \textit{hyper-representation} learning methods. 
Extensive empirical evaluation demonstrates that \ourmethod matches or exceeds state-of-the-art performance on several weight representation learning benchmarks, 
particularly in initialization for new tasks and larger ResNet architectures.\looseness-1 
\end{abstract}

%
%
\section{Introduction}
The exploration of the ``weight space'' of neural network (NN) models, i.e., the high-dimensional space spanned by the model parameters of a population of trained NNs, allows us to gain insights into the inner workings of those models. 

\begin{figure}[t!]
\centering
\includegraphics[width=1\columnwidth]{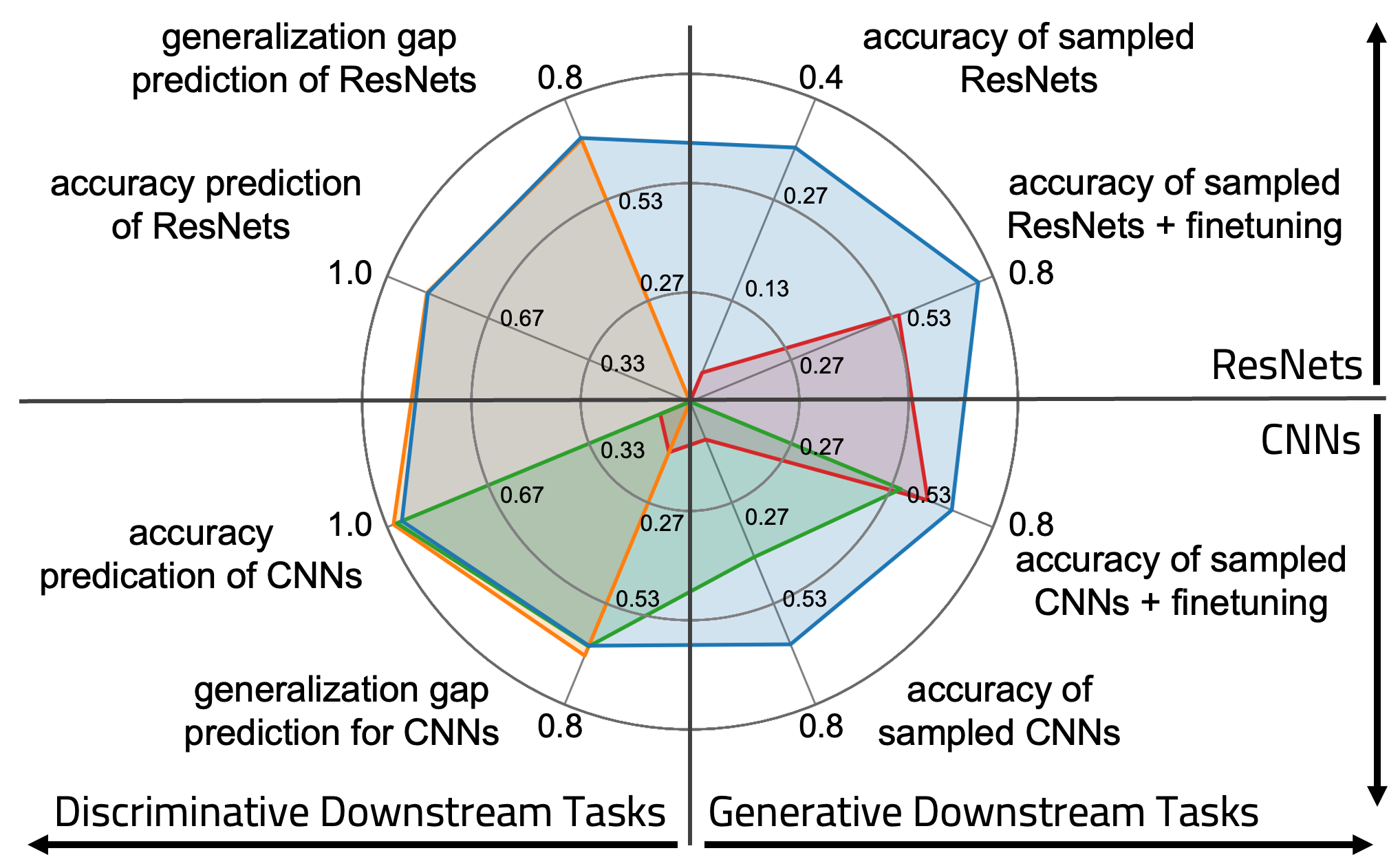}
\vspace{-0.7cm}
\caption{Aggregated results of 56 experiments showing \textbf{(left:)} four discriminative downstream tasks in $R^2$, and (\textbf{right:}) four generative downstream tasks in accuracy, each evaluated on (\textbf{bottom:}) CNNs model zoos trained on 4 datasets (NMIST, SVHN, CIFAR-10, STL) and (\textbf{top:}) ResNet18 model zoos trained on three datasets (CIFAR-10, CIFAR-100, Tiny-ImageNet).  
The colors indicate performance of {\color{Red}\textbf{\texttt{Red:}}} raw NN weights, {\color{RedOrange}\textbf{\texttt{Orange:}}} weight statistics from \citet{unterthinerPredictingNeuralNetwork2020}, {\color{Green}\textbf{\texttt{Green:}}} trained \textit{hyper-representations} from \citet{schurholtSelfSupervisedRepresentationLearning2021, schurholtHyperRepresentationsGenerativeModels2022}, and {\color{Blue}\textbf{\texttt{Blue:}}} \ourmethod (ours).
While some methods perform well on specific tasks, or are restricted by the size of the underlying models, \ourmethod can deliver excellent performance on all tasks and model sizes.
}
\vspace{-0.6cm}
\label{fig:radar_plot_overview}
\end{figure}

\vspace{0.1cm}

In the discriminative context, previous works aim to link weight space properties to properties such as model quality, generalization gap, or hyperparameters, using either the margin distribution \citep{yakTaskArchitectureIndependentGeneralization2019,jiangPredictingGeneralizationGap2019}, graph topology features \citep{corneanuComputingTestingError2020}, or eigenvalue decompositions of weight matrices \citep{martinTraditionalHeavyTailedSelf2019,MM20_SDM,martinImplicitSelfregularizationDeep2021,martinPredictingTrendsQuality2021}.
Some works learn classifiers to map between statistics of weights and model properties \citep{eilertsenClassifyingClassifierDissecting2020, unterthinerPredictingNeuralNetwork2020},
or learn lower-dimensional manifolds to infer NN model properties \citep{schurholtSelfSupervisedRepresentationLearning2021}.

\begin{figure*}[t!]
\centering
\includegraphics[width=0.75\paperwidth]{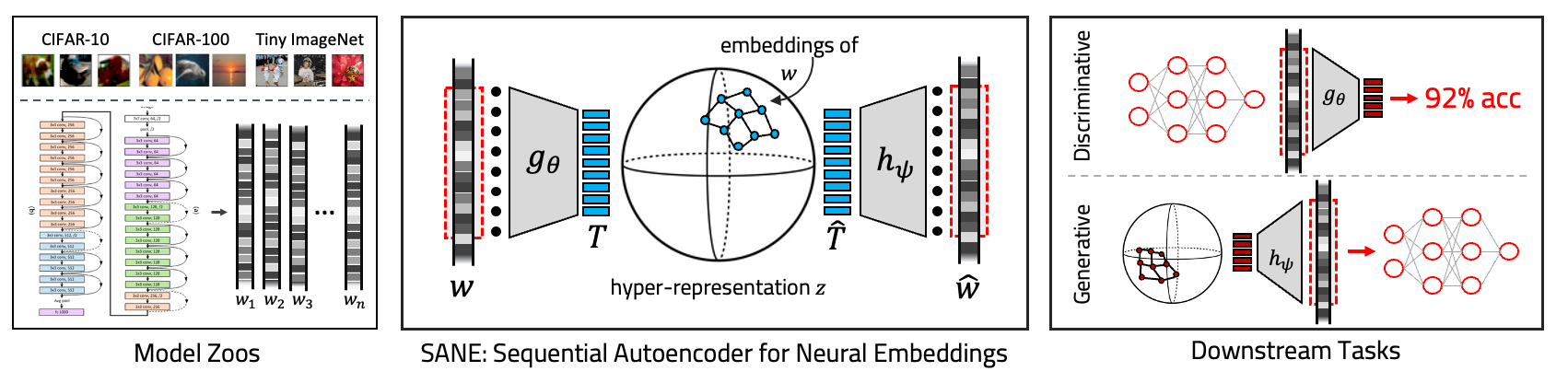}
\vspace{-0.55cm}
\caption{
Given model zoos trained on different classification tasks, we extract and sequentialize the model weights. \ourmethod trains \textit{hyper-representations} on 
weights subsequences, i.e., individual layers.
\ourmethod can be used for multiple downstream tasks, either using the encoder for discriminative tasks such as the prediction of model accuracy, or the decoder for generative tasks such as sampling of new models.
}
\vspace{-0.3cm}
\label{fig:approach_overview}
\end{figure*}

%
%
In the generative context, methods have been proposed to generate model weights using (Graph) HyperNetworks ~\citep{haHyperNetworks2016, zhangGraphHyperNetworksNeural2019, knyazevParameterPredictionUnseen2021}, Bayesian HyperNetworks \citep{deutschGeneratingNeuralNetworks2018}, HyperGANs \citep{ratzlaffHyperGANGenerativeModel2019}, and HyperTransformers \citep{zhmoginovHyperTransformerModelGeneration2022}.
These approaches have been used for tasks such as neural architecture search, model compression, ensembling, transfer learning, and meta-learning. They have in common that they derive their learning signal from the underlying (typically image) dataset. 
In contrast to these methods, 
so-called \textit{hyper-representations}~\citep{schurholtHyperRepresentationsGenerativeModels2022} learn a lower-dimensional representation directly from the weight space without the need to have access to data, e.g., the image dataset, to sample unseen NN models from that latent representation.

%
%
In this paper, we present \ourmethodlong (\ourmethod), an approach to learn task-agnostic representations of NN weight spaces capable of embedding individual NN models into a latent space to perform the above-mentioned discriminative or generative downstream tasks.
Our approach builds upon the idea of 
\textit{hyper-representations}~\citep{schurholtSelfSupervisedRepresentationLearning2021, schurholtHyperRepresentationsGenerativeModels2022}, which learn a lower-dimensional representation $z$ from a population of NN models. 
This is accomplished by auto-encoding their flattened weight vectors $w_i$ through a transformer architecture, with the bottleneck acting as a lower-dimensional embedding $z_i$ of each NN model. 
While the \textit{hyper-representation} method promises to be useful for discriminative and generative tasks, until now, separate \textit{hyper-representations} had to be trained specifically for either discriminative or generative tasks. 
Additionally, existing approaches have a major shortcoming: the underlying encoder-decoder model has to embed the entire flattened weight vectors $w_i$ at once into the learned lower-dimensional representation $z$. 
This drastically limits the size of NNs that can be embedded.
\ourmethod addresses these limitations by decomposing the entire weight vector $w_i$ into layers or smaller subsets, and then sequentially processes them. 
Instead of encoding the entire NN model by one embedding, \ourmethod encodes a potentially very large NN as multiple embeddings. 
The change from processing the entire flattened weight vector to subsets of weights is motivated by \citet{martinRethinkingGeneralizationRequires2019,martinImplicitSelfregularizationDeep2021}, who showed that global model information is preserved in the layer-wise components of NNs. 
An illustration of our approach can be found in Fig.~\ref{fig:approach_overview}.\looseness-1

To evaluate \ourmethod, we analyze how NN embeddings encoded by \ourmethod behave in comparison to \citet{martinRethinkingGeneralizationRequires2019,martinImplicitSelfregularizationDeep2021} quality measures. 
We show that some of these weight matrix quality metrics show similar characteristics as the embeddings produced by \ourmethod. 
This holds not only for held-out NN models of the model zoo used for training \ourmethod but also for NN models of out-of-distribution model zoos with different architectures and training data.
Further, we demonstrate that \ourmethod can learn \textit{hyper-representations} of much larger NN models, and so it makes them applicable to real-world problems. 
In particular, the models in the ResNet model zoo used for training are three orders of magnitudes larger than all model zoos used for \textit{hyper-representation} learning in previous works. 
While previous \textit{hyper-representation} learning methods were structurally constrained to encode the entire NN model at once, \ourmethod is scalable by applying its sequential approach to encode layers or subsets of weights into hyper-representation embeddings.
While we demonstrate scaling up to ResNet-101 models, \ourmethod is not fundamentally limited to that size.
Finally, we evaluate \ourmethod on both discriminative and generative downstream tasks.
For discriminate tasks, we evaluate on six model zoos by linear-probing for properties of the underlying NN models. 
For generative tasks, we evaluate on seven model zoos by sampling targeted model weights for initialization and transfer learning. 

We provide an aggregated overview of our results in Fig.~\ref{fig:radar_plot_overview}. 
On very small CNN models (evaluated on MNIST, SVHN, CIFAR-10, and STL, which we include for comparison with prior work), \ourmethod performs as well as previous state-of-the-art (SOTA) in discriminative tasks. In generative downstream tasks, \ourmethod outperforms SOTA by 25\% in accuracy for initialization on the same task and 17\% in accuracy for finetuning to new tasks.
On larger models such as ResNets (evaluated on CIFAR-10, CIFAR-100, Tiny-ImageNet, which were beyond the capabilities of prior work), we show results comparable to baselines for discriminative downstream tasks, and we report outperformance to baselines for generative downstream tasks by 31\% for initialization and 28\% for finetuning to new tasks. 
Additionally, we show that \ourmethod can sample targeted models by prompting with different architectures than it used for training. These sampled models can outperform models trained from scratch on the prompted architecture. Code is available at \href{https://github.com/HSG-AIML/SANE}{github.com/HSG-AIML/SANE}.\looseness-1

%
%
\section{Methods}

\textit{Hyper-representations} learn an encoder-decoder model on the weights of NNs \citep{schurholtSelfSupervisedRepresentationLearning2021}: 
    \begin{align}
        \mathbf{z} = g_{\theta}(\mathbf{W}) \label{eq:hrep_enc}\\
        \mathbf{\widehat{W}} = h_{\psi}( \mathbf{z} ),\label{eq:hrep_dec}
    \end{align}
where $g_{\theta}$ is the encoder which maps the flattened weights $\mathbf{W}$ to embeddings $\mathbf{z}$, and $h_{\psi}$ decodes back to reconstructed weights $\mathbf{\widehat{W}}$.
Even though previous work realized both encoder and decoder with transformer backbones, the weight vector had to be of fixed size, and models are represented in a global embedding space \citep{schurholtSelfSupervisedRepresentationLearning2021, schurholtHyperRepresentationsGenerativeModels2022}. 
\textit{Hyper-representations} are trained with a reconstruction loss $\mathcal{L}_{rec}=\|\mathbf{W}-\mathbf{\widehat{W}} \|_2^2$ and contrastive guidance loss $\mathcal{L}_{c} = NTXent(p_{\phi}(\mathbf{z_i}),p_{\phi}(\mathbf{z_j}))$, where $p_{\phi}$ is a projection head. \citet{schurholtSelfSupervisedRepresentationLearning2021} proposed weight permutation, noise, and masking as augmentations to generate views $i,j$ of the same model.

Existing \textit{hyper-representation} methods have two major limitations: 
i) using the full weight vector to compute global model embeddings becomes infeasible for larger models; 
and ii) they can only embed models that share the architecture with the original model zoo.
Our \ourmethod method addresses both of these limitations. 
To make models more digestible for pretraining and inference, we propose to express models as sequences of token vectors.
To address i), \ourmethod learns per-token embeddings, which are trained on subsequences of the full base model sequence. 
This way, the memory and compute load are decoupled from the base model size. 
By decoupling the tokenization from the representation learning, we also address ii). 
The models in the model zoo set can have varying architectures, as long as they are expressed as a sequence with the same token-vector size. 
The transformer backbone and per-token embeddings also allow changes to the length of the sequence during or after training. 
Below, we provide technical details on \ourmethod.
We first provide details on pretraining \ourmethod, computing model embeddings, and sampling models; and we then introduce additional \textit{aligning}, \textit{haloing}, and \textit{bn-conditioning} methods to stabilize training and inference.

\textbf{\ourmethod: \ourmethodlong}
\label{sec:seq_hyper_reps}
To tokenize weights, we reshape the weights  $\mathbf{W}_{raw} \in \mathbb{R}^{c_{out}\times c_{1}\times\cdots \times c_{in}}$ to 2d matrices $\mathbf{W} \in \mathbb{R}^{c_{out}\times c_{r}}$, where $c_{out}$ are the outgoing channels, and where $c_{r}$ the remaining, flattened dimensions. 
We then slice the weights row-wise, along the outgoing channel. 
Using global token size $d_t$, we split the slices into multiple parts if $c_r>d_t$ and zero-pad to fill up to $d_t$. 
For weights $\mathbf{W}_l$ of layer $l$, this gives us tokens $\mathbf{T}_l \in \mathbb{R}^{n_l \times d_t}$, where $n_l = c_{out,l} \, \mbox{ceil}(\frac{c_r}{d_t})$. 
Since all tokens $\mathbf{T}_l$ share the same token size, the tokens of layer $l=1,...,L$ can be concatenated to get the model token sequence $\mathbf{T} \in \mathbb{R}^{N\times d_t}$.
To indicate the position of a token, we use a 3-dimensional position $\mathbf{P}_n = [n,l,k]$, where $n\in[1, N]$ indicates the global position in the sequence, $l \in [1, L]$ indicate the layer index, and $k \in [1, K(l)]$ is the position of the token within the layer.

Out of the full token sequence $\mathbf{T}$ and positions $\mathbf{P} \in \mathbb{N}^{N \times 3}$, we take a random consecutive sub-sequence $\mathbf{T}_{s,n}=\mathbf{T}_{n,...,n+ws}$ with positions $\mathbf{P}_{s,n}=\mathbf{P}_{n,...,n+ws}$ of length $ws$. 
We call these sub-sequences windows and the length of the sub-sequence the window size $ws$.

For \ourmethod on windows of tokens, we extend Eqs. \ref{eq:hrep_enc} and \ref{eq:hrep_dec} to encode and decode token windows as
    \begin{align}
        \mathbf{z}_{s,n} &= g_{\theta}(\mathbf{T}_{s,n},\mathbf{P}_{s,n}) \label{eq:seqhrep_enc}\\
        \widehat{\mathbf{T}}_{s,n} &= h_{\psi}( \mathbf{z}_{s,n},\mathbf{P}_{s,n}),\label{eq:seqhrep_dec}
    \end{align}
where $\mathbf{z}_{s,n} \in \mathbb{R}^{ws \times d_z}$ is the per-token latent representation of the window. 
In contrast to \textit{hyper-representations} Eqs. \ref{eq:hrep_enc} and \ref{eq:hrep_dec} which operate on the full flattened weights of a model, \ourmethod encodes sub-sequences of tokenized models.
For simplicity, we apply linear mapping to and from the bottleneck, to reduce tokens from $d_t$ to $d_z$. 

We adapt the composite training loss of \textit{hyper-representations}, $\mathcal{L} = (1-\gamma) \mathcal{L}_{rec} + \gamma \mathcal{L}_c$, for sequences as:
\begin{align}
    \mathcal{L}_{rec} &=\| \mathbf{M}_{s,n} \odot \left( \mathbf{T}_{s,n}-\widehat{\mathbf{T}}_{s,n} \right)\|_2^2 \label{eq:seqhrep_recon_loss} \\
    \mathcal{L}_{c} &= NTXent(p_{\phi}(\mathbf{z_{s,n,i}}),p_{\phi}(\mathbf{z_{s,n,j}})). \label{eq:seqhrep_con_loss}
\end{align}
Here, the mask $\mathbf{M}_{s,n}$ indicates signal with $1$ and padding with $0$, to ensure that the loss is only computed on actual weights. The contrastive guidance loss uses the augmented views $i,j$ and projection head $p_{\phi}$.

The pretraining procedure is detailed in Algorithm \ref{alg:pretraining}. 
We preprocess model weights by standardizing weights per layer and aligning all models to a reference model; see \textit{Model Alignment} below.
As in previous work \citep{schurholtSelfSupervisedRepresentationLearning2021,peeblesLearningLearnGenerative2022}, 
the encoder and decoder are realized as transformer blocks. 
Training on the full sequence would memory-limit the base-model size by its sequence length.
\begin{algorithm}[]
   \caption{\ourmethod pretraining}
   \label{alg:pretraining}
\begin{algorithmic}
    \STATE {\bfseries Input:} population of models
    \STATE \textbf{i:} standardize models weights
    \STATE \textbf{ii:} align models to one common reference model
    \STATE \textbf{iii:} tokenize models to tokens $\mathbf{T}$, positions $\mathbf{P}$, masks $\mathbf{M}$
    \STATE \textbf{iv:} draw \textit{k} windows per model: $\mathbf{T}_{s,n}$, $\mathbf{P}_{s,n}$,  $\mathbf{M}_{s,n}$
    \STATE \textbf{v:} train on $\mathcal{L}_{train}$ until convergence of $\mathcal{L}_{val}$
\end{algorithmic}
\end{algorithm}
Training the encoder and decoder on windows instead of the full model sequence decouples the memory requirement from the base model's full sequence length. 
The window size can be used to balance GPU memory load and the amount of context information.
Notably, since we disentangle the tokenization from the representation learning model, \ourmethod also allows us to embed sequences of models with varying architectures, as long as their token size is the same. 
To prevent potential overfitting to specific window positions, we propose to sample windows from each model sequence multiple times randomly.

\textbf{Computing \ourmethod Model Embeddings.}
\ourmethod can be used to analyze models in embedding space, e.g., by using embeddings as features to predict properties such as accuracy or to identify other model quality metrics.
In contrast to \textit{hyper-representations}, \ourmethod can embed different model sizes and architectures in the same embedding space.
To embed any model, we begin by preprocessing weights by standardizing per layer and aligning models to a pre-defined reference model (see \textit{Model Alignment} below).
Subsequently, the preprocessed models are tokenized as described above.
For short model sequences, the embedding sequences can be directly computed as $\mathbf{z} = g_\theta(\mathbf{T},\mathbf{P})$. For larger models, the token sequences are too long to embed as one. We therefore employ \textit{haloing} (see below) to encode the entire sequence as coherent subsequences.
Algorithm \ref{alg:embeddings} summarizes the embedding computation. 
\vspace{-4pt}
\begin{algorithm}[]
   \caption{\ourmethod model embedding computation}
   \label{alg:embeddings}
\begin{algorithmic}
    \STATE {\bfseries Input:} population of models
    \STATE \textbf{i:} preprocessing: standardize and align model weights
    \STATE \textbf{ii:} tokenize models: $\mathbf{T}$, positions $\mathbf{P}$, property $y$
    \STATE \textbf{iii:} split $\mathbf{T}$, $\mathbf{P}$ to consecutive chunks $\mathbf{T}_{hs,n}, \mathbf{P}_{hs,n}$ 
    \STATE \textbf{iv:} compute embeddings $\mathbf{z}_{hs,n} = g_{\theta}(\mathbf{T}_{hs,n},\mathbf{P}_{hs,n})$
    \STATE \textbf{v:} stitch model embeddings $\mathbf{z}$ together from chunks $\mathbf{z}_{hs,n}$
\end{algorithmic}
\end{algorithm}
\vspace{-8pt}
To compare different models in embedding space, we aggregate the sequences of token embeddings. To that end, we understand the token sequence of one model to form a surface in embedding space and choose to represent that surface by its center of gravity. That is, we take the mean of all tokens along the embedding dimension as $\mathbf{\bar{z}} = \frac{1}{N} \sum_{n=1}^N(\mathbf{z}_n)$. That results in one vector in embedding space per model. Of course, one could use other aggregation methods with \ourmethod.

\textbf{Sampling Models with Few Prompt Examples.}
Sampling models from \ourmethod promises to transfer knowledge from existing populations to new models with different architectures.
Given pretrained encoders $g_{\theta}$ and decoder $h_\psi$, the challenge is to identify the distribution $\mathcal{P}$ in latent space which contains the targeted properties.
To approximate that distribution, previous work used a large number of well-trained models \citep{peeblesLearningLearnGenerative2022,schurholtHyperRepresentationsGenerativeModels2022}.
However, increasing the size of the sampled models makes generating a large number of high-performance models exceedingly expensive.
Instead of using expensive high-performance models to model $\mathcal{P}$ directly, we propose to find a rough estimate of $\mathcal{P}$, sample broadly, and refine $\mathcal{P}$ using the signal from the sampled models.
Using $E$ prompt examples $\mathbf{W}^e$ we compute the token sequence $\mathbf{T}^e$ and corresponding embedding sequence $\mathbf{z}^e = g_{\theta}(\mathbf{T}^e, \mathbf{P})$. Following previous work, we model the distribution $\mathcal{P}$ with a Kernel Density Estimation (KDE) per token as $\mathcal{P}_{e \in E}(\mathbf{z}_n^e)$ ~\citep{schurholtHyperRepresentationsGenerativeModels2022}.
We then draw $k$ new token samples as:
\begin{equation}
    \mathbf{z}_n^k \sim \mathcal{P}_{e \in E}(\mathbf{z}_n^e).
\end{equation}
We reconstruct the sampled embeddings to weight tokens $\mathbf{T}^k = h_{\psi}(\mathbf{z}^k,\mathbf{P})$ and then weights $\mathbf{W}^k$. 
Sampling tokens can be done cheaply, decoding and evaluating the weights using some performance metric involves only forward passes and is likewise cheap. 
Therefore, one can draw a large amount of samples and keep only the top $m$ models, according to the performance metric. 
We call this method \emph{subsampling}.
The process can be refined iteratively, by re-using the embeddings $\mathbf{z}^k$ of the best models as new prompt examples, to adjust the sampling distribution to best fit the needs of the performance metric. 
We call this sampling method \emph{bootstrapped}.
By only requiring a rough version of $\mathcal{P}$ and refining with the target signal, our sampling strategy reduces requirements on prompt examples such that only very few and slightly trained prompt examples are necessary. The overall sampling method is outlined in Algorithm~\ref{alg:generating}. It makes use of \textit{model alignment}, \textit{haloing}, and \textit{batch-norm conditioning} which are detailed below.
In addition to the compute efficiency, these sampling methods learn the distribution of targeted models in embedding space. 
Further, they are not bound to the distribution of prompt examples, but instead they can find the distribution that best satisfies the target performance metric, independent of the prompt examples.
\begin{algorithm}[]
   \caption{Sampling models with \ourmethod }
   \label{alg:generating}
\begin{algorithmic}
    \STATE {\bfseries Input:} model prompt examples $\mathbf{W}^e$
    \STATE \textbf{i:} tokenize prompt examples: tokens $\mathbf{T}^e$, positions $\mathbf{P}^e$
    \STATE \textbf{ii:} embed prompt examples $\mathbf{z}^e$ following Alg. \ref{alg:embeddings}
    \FOR{$i_{boot} = 1$ {\bfseries to} \textit{bootstrap iterations}}
        \STATE \textbf{iii:} draw $k$ samples $\mathbf{z}_n^k \sim \mathcal{P}_{e \in E}(\mathbf{z}_n^e)$
        \STATE \textbf{iv:} decode to tokens $\mathbf{T^k} = h_{\psi}( \mathbf{z}^k )$
        \STATE \textbf{v:} apply batch-norm conditioning
        \STATE \textbf{vi:} compute target metric and keep best $m$ models
        \IF{\textit{bootstrap iterations}$\; > 1$}
            \STATE \textbf{vii:} $\mathbf{z}^e = \mathbf{z}^k for \; k \in m$ 
        \ENDIF
    \ENDFOR
\end{algorithmic}
\end{algorithm}
\vspace{-8pt}

Growing sample model size poses several additional challenges, three of which we address with the following methods. We evaluate these methods in Appendix \ref{sec:ablation}.\looseness-1

\textbf{Model Alignment.}
Symmetries in the weight space of NN complicate representation learning of the weights. The number of symmetries grows fast with model size \citep{bishopPatternRecognitionMachine2006}. 
To make representation learning easier, we reduced all training models to a unique, canonical basis of a reference model.
With reference model $A$ we align model $B$ by finding the permutation $\pi = argmin_{\pi} \|\text{vec}(\Theta{(A)}) - \text{vec}(\Theta(B))\|^2$, where $\Theta(A)$ are the parameters of model $A$ \citep{ainsworthGitReBasinMerging2022}. 
We fix the same reference model across all dataset splits and use the last epoch of each model to determine the permutation for that model.

\textbf{Haloing.}
The sequential decomposition of \ourmethod decouples the pretraining sequence length from downstream task sequence lengths.
Since the memory load at inference is considerably lower, the sequences at inference can be longer. 
However, full model sequences may still not fit in memory and may have to be processed in slices.
To ensure consistency between the slices, we add context around the content windows. With added context halo before and after the content window, we get
$\mathbf{T}_{hs,n}=\mathbf{T}_{n-h,..,n,...,n+ws,n+ws+h}$. 
Similar to approaches in computer vision \citep{vaswaniScalingLocalSelfAttention2021}, this context halo is added for the pass through encoder and decoder, but disregarded after.

\textbf{Batch-Norm Conditioning.}
In most current NN models, some parameters like batch-norm weights are updated during forward passes instead of with gradients. 
Since that makes them structurally different, we exclude these parameters from representation learning and sampling with \ourmethod.
Nonetheless, these parameters need to be instantiated for sampled models to work well. 
For model sampling methods, we therefore propose to condition batch-norm parameters by performing a few forward passes with some target data. 
Importantly, this process does not update the learned weights of the model. 
It serves to align the batch norm statistics with the model's weights.

%
%
\section{Training \ourmethod}
We pretrain \ourmethod following Alg. \ref{alg:pretraining} on several populations of trained NN models, from the model zoo dataset \citep{schurholtModelZoosDataset2022}. 
We use zoos of small models to compare with previous work, as well as zoos containing larger ResNet-18 models. 
All zoos are split into training, validation, and test splits $70:15:15$.
\vspace{-10pt}
\begin{itemize}
\item \textbf{Smaller CNN zoos.}
The MNIST and SVHN zoos contain LeNet-style models with 3 convolution and 2 dense layers and only $\sim 2.5k$ parameters. 
The slightly larger CIFAR-10 and STL-10 zoos use the same architecture with wider layers and $\sim 12k$ parameters.
\vspace{-6pt}
\item \textbf{Larger ResNet zoos.}
We also use the CIFAR-10, CIFAR-100, and Tiny-Imagenet zoos containing ResNet-18 models \citep{schurholtModelZoosDataset2022} with $\sim 12M$ parameters to evaluate scalability to large models. 
\end{itemize}

\vspace{-5pt}
\textbf{Pretraining.}
We train \ourmethod using Alg. \ref{alg:pretraining}.
As augmentations, we use noise and permutation. 
The permutation is computed relative to the aligned model. For contrastive learning, the aligned model serves as one view, and a permuted version as the second view. \looseness-1

\textbf{Implementation Details.} 
To maintain diversity within each batch, we select only a single window from each model. Loading, preprocessing, and augmenting the entire sample, only to use ca. 1\% of it, is infeasible. To address this, we leverage FFCV \citep{leclercFFCVAcceleratingTraining2023} to compile datasets consisting of sliced and permuted windows of models. Each model is super-sampled for approximately full coverage within the training set, considering the ratio of window length to sequence length.
For the ResNet zoos, we include 140 models per zoo, a number that remains manageable in terms of memory and storage. 
We train for 50 epochs using a OneCycle learning rate scheduler \citep{smithSuperConvergenceVeryFast2018}. Seeds are recorded to ensure reproducibility.
We build \ourmethod in PyTorch \citep{paszkePyTorchImperativeStyle2019}, using automatic mixed precision and flash attention \citep{daoFlashAttentionFastMemoryEfficient2022} to enhance performance. We use ray.tune \citep{liawTuneResearchPlatform2018} for hyperparameter optimization.

\vspace{-4pt}
\section{Embedding Analysis}
\vspace{-2pt}
In this section, we analyze the embeddings of \ourmethod and compare to the weight-analysis methods \textit{WeightWatcher} (WW) \citep{martinPredictingTrendsQuality2021}. 
We focus on three aspects: 
i) global relation between accuracy and embeddings; 
ii) the trend of embeddings over layer index, as in  \citep{martinPredictingTrendsQuality2021}; and
iii) the identification of training phases as in \citep{martinTraditionalHeavyTailedSelf2019,martinImplicitSelfregularizationDeep2021}.
\vspace{-2pt}

To analyze weights, we focus on two WW metrics which in previous work reveal model performance as well as internal model composition (\textit{correlation flow}); the log spectral norm $\log(\| \mathbf{W}\|^2_{\infty})$ and weighted $\alpha$, the coefficient of the power law fitted to the empirical spectral density \citep{martinPredictingTrendsQuality2021}. 
These two metrics describe different aspects of the eigenvalue distribution. 
To get a similar signal on the internal dependency of weight matrices, we compute per-layer scalars $\hat{z}_l$ as the spread of the tokens of one layer in hyper-representation space, i.e., their standard deviation: 
\begin{align}
\vspace{-4pt}
    \hat{z}_l &=  std_t(\textbf{z}^t_m) \\
    \textbf{z}^t_m &= g(\textbf{W}^t_m),
\vspace{-4pt}
\end{align}
where $g$ is the hyper-rep encoder, $\textbf{z}^t_m$ are the stacked tokens $t$ of layer $m$, and $\textbf{W}^t_m$ is the weight-slice $t$ of layer $m$.
\vspace{-2pt}

\begin{figure}[h]
\centering
\includegraphics[trim=4mm 0mm 1.5mm 1.5mm, width=1.0\linewidth]{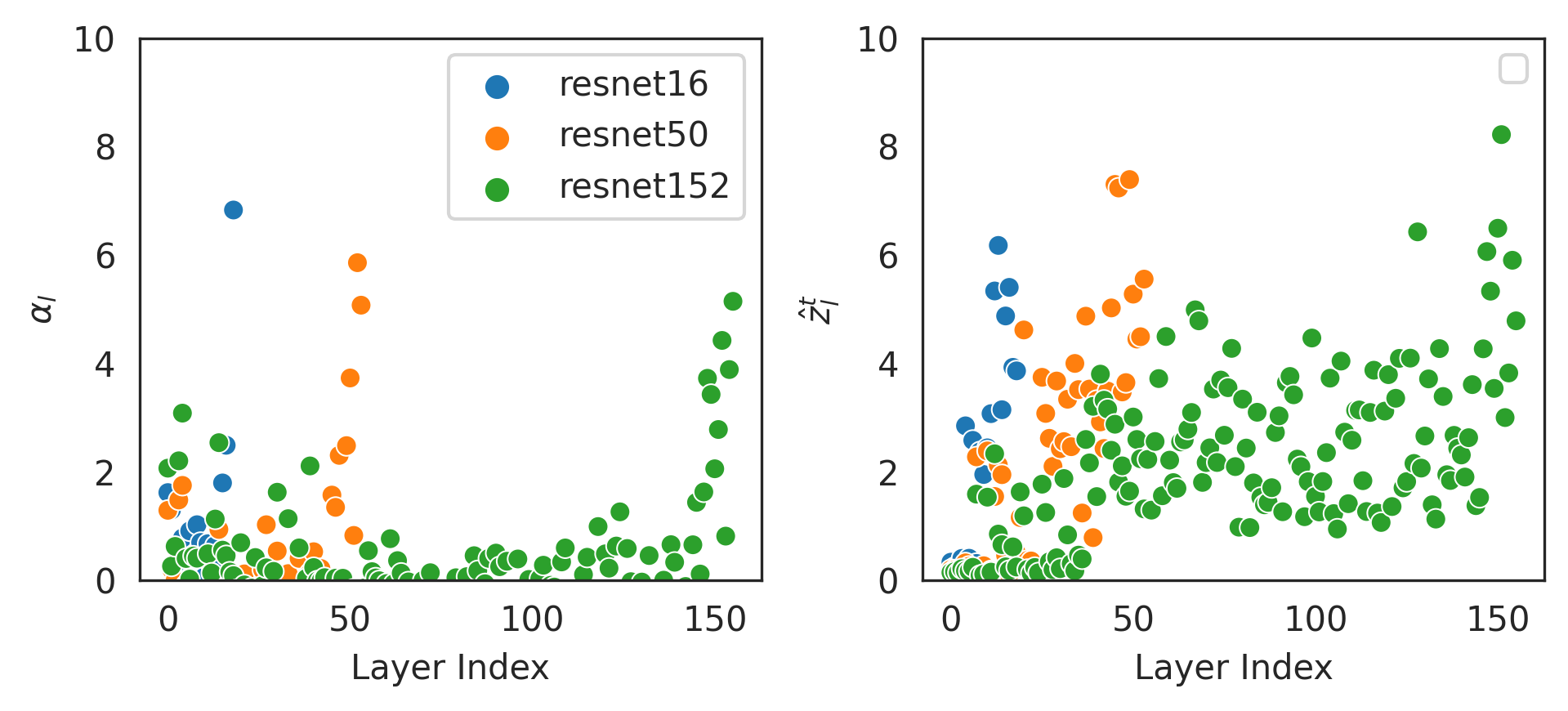}
\vspace{-20pt}
\captionof{figure}{Comparison between WeightWatcher (WW) features (left) and \ourmethod (right). Features over layer index for ResNets from pytorchcv of different sizes.  
}
\vspace{-2mm}
\label{fig:ww_comparison_layers_resnet}
\end{figure}

To compare WW metrics to \ourmethod,
we pretrain \ourmethod on a Tiny-Imagenet ResNet-18 zoo and compute the two metrics on ResNets and VGGs of different sizes trained on ImageNet from pytorchcv \citep{semeryOsmrImgclsmob2024}.
On both ResNets in Figure \ref{fig:ww_comparison_layers_resnet}, \ref{fig:ww_comparison_layers_resnets_2x2} and VGGs in Figure \ref{fig:ww_comparison_layers_vgg}, the WW metrics and our embeddings show similar global trends. 
On ResNets, our embeddings and WW have low values at early layers and a sharp increase at the end. 
However, our embeddings add an additional step for intermediate layers, which may indicate that \ourmethod is sensitive to a higher degree of variation in these layers which previous work found by comparing activations \citep{kornblithSimilarityNeuralNetwork2019}.\looseness-1 

\begin{figure}[h]
\centering
\includegraphics[trim={1mm 0mm 1.5mm 5.5mm}, clip, width=0.85\linewidth]{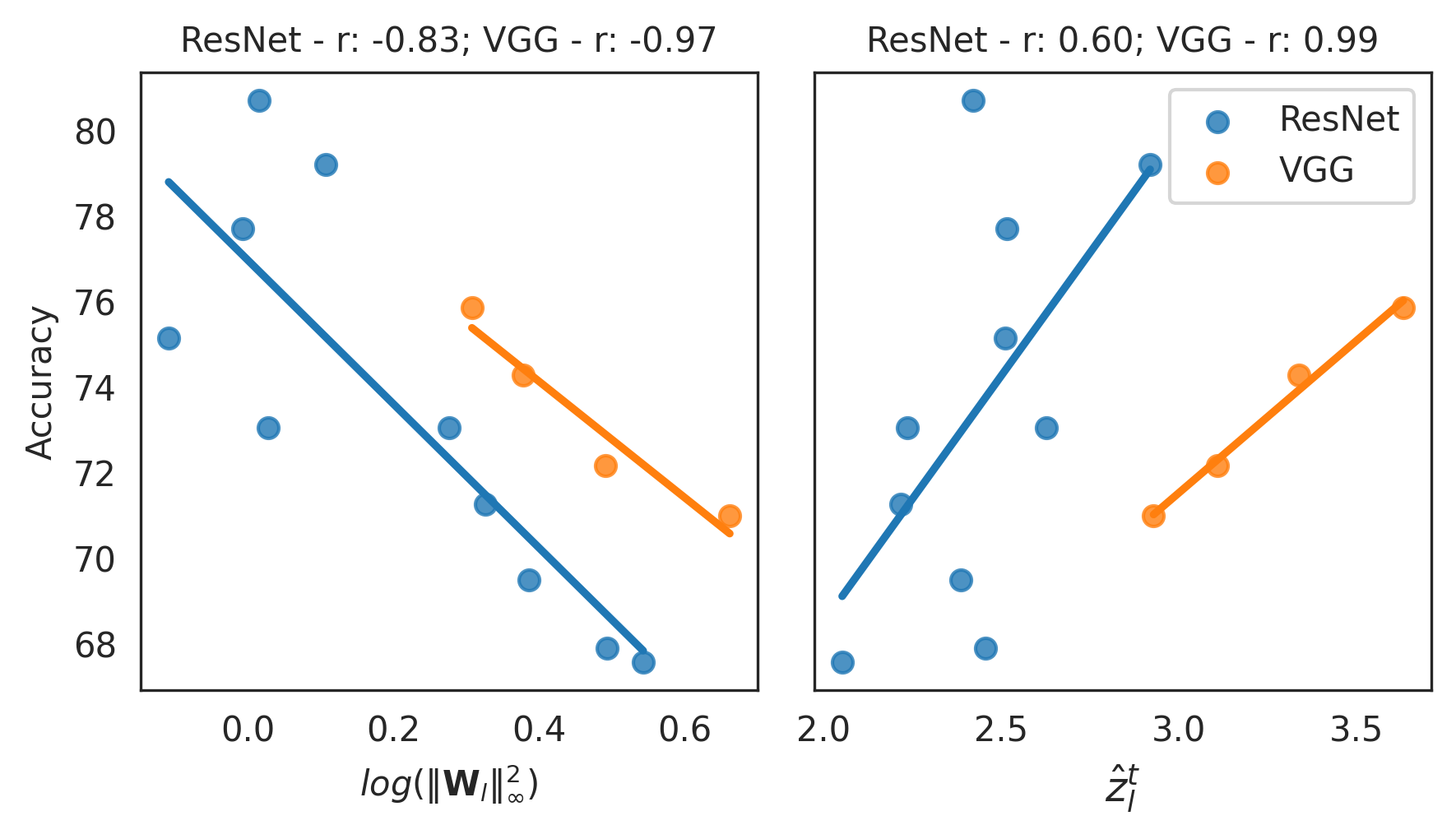}
\vspace{-14pt}
\captionof{figure}{Comparison between WeightWatcher features (left) and \ourmethod (right). Accuracy over model features for ResNets and VGGs from pytorchcv of different sizes. Although \ourmethod is pretrained in a self-supervised fashion, it preserves the linear relation of a globally-aggregated embedding to model accuracy.
}
\vspace{-3mm}
\label{fig:ww_comparison_accuracy}    
\end{figure}

In a second experiment, we aggregate the layer-wise embeddings $\hat{z}_l$ to evaluate relations to model accuracy in Figures \ref{fig:ww_comparison_accuracy}, \ref{fig:ww_comparison_accuracy_2x2} and \ref{fig:ww_comparison_accuracy_our_zoo}, similar to previous work \citep{martinPredictingTrendsQuality2021}. 
On models from pytorchcv and the Tiny-ImagNet model zoo from \citep{schurholtModelZoosDataset2022}, the WW features and \ourmethod embeddings both show strong correlations to model accuracy. 
However, while the WW metrics are negatively correlated to accuracy, our embeddings are positively correlated to accuracy. 
The reason for that may lie in the additional 'step' in Figure \ref{fig:ww_comparison_layers_resnet}. 
That is, larger models with more layers generally have higher performance. As Figure \ref{fig:ww_comparison_layers_resnet} shows, more layers add very small values reducing the global average for WW metrics. For our embeddings, deeper models have more layers with higher $\hat{z}_l$ values, due to the afore-mentioned step. This increases the global model average with growing model size.
Lastly, we compare the eigenvalue spectrum to embeddings. Previous work identified distinct shapes at different training phases or with varying training hyperparameters  \citep{martinTraditionalHeavyTailedSelf2019,martinImplicitSelfregularizationDeep2021}. While we can replicate the distributions of the eigenvalues, the distributions of our embeddings only show the change from early phases of training to the heavy-tailed distribution; see Figure \ref{fig:ww_comparison_phases}. 

In summary, our embedding analysis indicates that \ourmethod represents several aspects of model quality (globally and on a layer level) that have been established previously.
%

\section{Empirical Performance}
In this section, we describe the general performance \ourmethod.
\subsection{Predicting Model Properties}
\label{sec:property_prediction}
We evaluate \ourmethod for discriminative downstream tasks as a proxy for encoded model qualities. Specifically, we investigate whether \ourmethod matches the predictive performance of \textit{hyper-representations} on small CNN models (Table \ref{tab:discr_small_zoos}) and whether similar performance can be achieved on ResNet-18 models (Table \ref{tab:discr_resnet_zoos}). 
To that end, we compute model embeddings $\mathbf{\bar{z}}$ as outlined in Alg. \ref{alg:embeddings}, and we compare against flattened weights $W$ and weight statistics $s(W)$.
Following the experimental setup of \citep{eilertsenClassifyingClassifierDissecting2020,unterthinerPredictingNeuralNetwork2020,schurholtSelfSupervisedRepresentationLearning2021}, we compute embeddings using the three methods and linear probe for test accuracy (\texttt{Acc}), epoch (\texttt{Ep}), and generalization gap (\texttt{Ggap}). We again use trained models from the modelzoo repository \citep{schurholtModelZoosDataset2022}, with the same train, test, val splits as above. 
\begin{table}[h]
\vspace{-4mm}
\captionof{table}{
Property prediction on populations of small CNNs used in previous work \citep{schurholtSelfSupervisedRepresentationLearning2021}. We report the regression $R^2$ on the test set prediction test accuracy \texttt{Acc.}, epoch \texttt{Ep.} and generalization gap \texttt{Ggap} for linear probing with model weights $W$, model weights statistics $s(W)$ or \ourmethod embeddings as inputs.\looseness-1}
\label{tab:discr_small_zoos}
\centering
\sc
\scriptsize
\setlength{\tabcolsep}{2.5pt}
\begin{tabularx}{1.0\linewidth}{lccccccccccc}
\toprule
      & \multicolumn{3}{c}{MNIST}                                                      &  & \multicolumn{3}{c}{SVHN}                                                       &  & \multicolumn{3}{c}{CIFAR-10 (CNN)}                                             \\ 
      \cmidrule(r){2-4} \cmidrule(lr){6-8} \cmidrule(lr){10-12}
      & \multicolumn{1}{c}{W} & \multicolumn{1}{c}{s(W)} & \multicolumn{1}{c}{\ourmethod} &  & \multicolumn{1}{c}{W} & \multicolumn{1}{c}{s(W)} & \multicolumn{1}{c}{\ourmethod} &  & \multicolumn{1}{c}{W} & \multicolumn{1}{c}{s(W)} & \multicolumn{1}{c}{\ourmethod}  \\ 
      \cmidrule(r){1-1} \cmidrule(r){2-4} \cmidrule(lr){6-8} \cmidrule(lr){10-12}
Acc.   & 0.965                 & \textbf{0.987}           & 0.978                       &  & 0.910                 & 0.985                    & \textbf{0.991}              &  & -7.580                & \textbf{0.965}           & 0.885                        \\
Ep. & 0.953                 & \textbf{0.974}           & 0.958                       &  & 0.833                 & \textbf{0.953}           & 0.930                       &  & 0.636                 & \textbf{0.923}           & 0.771                       \\
Ggap  & 0.246                 & 0.393                    & \textbf{0.402}              &  & 0.479                 & 0.711                    & \textbf{0.760}              &  & 0.324                 & \textbf{0.909}           & 0.772                       \\ 
\bottomrule
\end{tabularx}
\vspace{-2mm}
\end{table} 

\textbf{\ourmethod matches baselines on small models.}
The results of linear probing on small CNNs in Tables \ref{tab:discr_small_zoos} and \ref{tab:discr_small_zoos_full} confirm the performance of $W$ (low) and $s(W)$ (very high) of previous work. 
\ourmethod embeddings show comparably high performance to the $s(W)$ and previous \textit{hyper-representations}. 
Additional experiments in Appendix \ref{app:discr_comparison_previous_work} compare to previous work and confirm these findings. Sequential decomposition and representation learning as well as using the center of gravity does not significantly reduce the information contained in \ourmethod embeddings. 
%
\begin{table}[h]
\vspace{-2mm}
\captionof{table}{
Property prediction on ResNet-18 model zoos of \citep{schurholtModelZoosDataset2022}. We report the regression $R^2$ on the test set prediction test accuracy \texttt{Acc.}, epoch \texttt{Ep.} and generalization gap \texttt{Ggap} for linear probing with model weights statistics $s(W)$ or \ourmethod embeddings as inputs.\looseness-1
}
\label{tab:discr_resnet_zoos}
\centering
\sc
\footnotesize
\setlength{\tabcolsep}{3.5pt}
\begin{tabularx}{1.0\linewidth}{lcccccccc}
\toprule
      & \multicolumn{2}{c}{CIFAR-10} &  & \multicolumn{2}{c}{CIFAR-100} &  & \multicolumn{2}{c}{Tiny-Imagenet} \\
      \cmidrule(){2-3} \cmidrule(){5-6} \cmidrule(){8-9} 
      &  s(W)     & \ourmethod    &  & s(W)     & \ourmethod    &      & s(W)      & \ourmethod      \\ 
\cmidrule(r){1-1} \cmidrule(){2-3} \cmidrule(){5-6} \cmidrule(){8-9} 
Acc.      & 0.880    & 0.879      &       & 0.923    & 0.922      &        &  0.802    &       0.795       \\
Ep.       & 0.999    & 0.999      &       & 0.999    & 0.992      &        &  0.999    &       0.980       \\
Ggap      & 0.490    & 0.512      &       & 0.882    & 0.879      &        &  0.704    &       0.699 
\\
\bottomrule
\end{tabularx}
\vspace{-4mm}
\end{table}

\newpage
\textbf{\ourmethod performance prediction scales to ResNets.} 
Both $s(W)$ and \ourmethod embeddings show similarly high performance on populations of ResNet-18s; see Table \ref{tab:discr_resnet_zoos}. 
On ResNet-18s, using the full weights $W$ for linear probing is infeasible due to the size of the flattened weights.
\ourmethod matches the high performance of $s(W)$. 
The results show that sequential \textit{hyper-representations} are capable of scaling to ResNet-18 models. 
Further, the aggregation even of long sequences (ca. 50k tokens) embedded in \ourmethod preserves meaningful information on model performance, which indicates the feasibility of applications like model diagnostics or targeted sampling.~\looseness-1
%
%
%
\subsection{Generating Models}
\label{sec:generating_models}
\begin{figure}[t!]
\centering
\includegraphics[trim=0mm 0mm 0mm 0mm, width=1.0\linewidth]{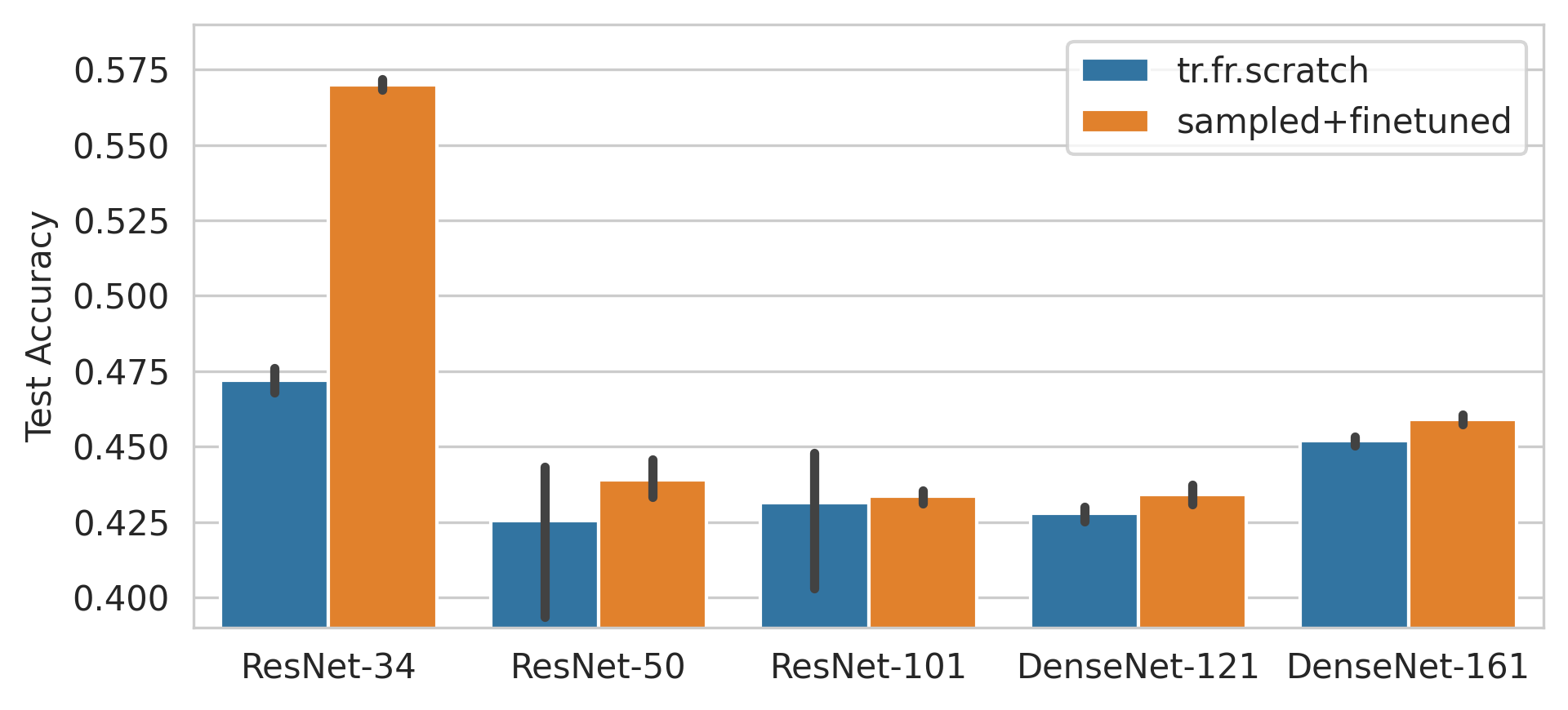}
\vspace{-2mm}
\captionof{figure}{Comparison between sampled models and random initialization trained for 5 epochs Tiny-Imagenet. Different architectures are sampled from \ourmethod pretrained on a ResNet-18 CIFAR-100 zoo. Although both models and tasks are changed, sampled models perform better.}
\label{fig:sampling_new_arch_task}    
\end{figure}

We evaluate \ourmethod for the generative downstream tasks i.e., for sampling model weights. 
We generate weights following Alg. \ref{alg:generating} and test them in fine-tuning, transfer learning, and how they generalize to new tasks and architectures. 
In the following paragraphs, we begin with experiments on small CNN models from the modelzoo repository to compare with previous work (Tables \ref{tab:generative_cnn_zoos_in_domain}, \ref{tab:generative_cnn_zoos_transfer}). 
Subsequently, we evaluate \ourmethod for sampling ResNet-18 models for finetuning and transfer learning (Tables \ref{tab:generative_resnets_zoos_in_domain}, \ref{tab:generative_resnets_zoos_transfer}). 
Lastly, we evaluate sampling for new tasks and new architectures using only few prompt examples (Figure \ref{fig:sampling_new_arch_task} and Tables \ref{tab:generative_resnets_zoos_fewshot_transfer}, \ref{tab:generative_resnets_zoos_fewshot_resnet34}, \ref{tab:generative_resnets_zoos_fewshot_ti_resnet34}, \ref{tab:generative_resnets_zoos_fewshot_resnet34_one_prompt_example}).

We pretrain \ourmethod on models from the first half of the training epochs with Alg. \ref{alg:pretraining}, and keep the remaining epochs (26-50) as holdout to compare against, following the experimental setup of \citep{schurholtHyperRepresentationsGenerativeModels2022}. 
We sample using Alg. \ref{alg:generating} and use models from the last epoch in the pretraining set (epoch 25) as prompt examples. 
We denote subsampling with \ourmethod$_{SUB}$ and iteratively updating the distribution $\mathcal{P}$ as \ourmethod$_{BOOT}$. To evaluate the impact of the sampling method, we also combine \ourmethod with the $KDE30$ sampling approach that uses high-quality prompt examples \citep{schurholtHyperRepresentationsGenerativeModels2022}. We further evaluate sampling without prompt examples by bootstrapping off of a Gaussian prior $\mathcal{P}$, denoted as \ourmethod$_{GAUSS}$. We compare against training from scratch, as well as fine-tuning from the prompt examples.


\begin{table}[t]
\vspace{-2mm}
\captionof{table}{
Model generation on CNN model populations fine-tuned on the same task. We compare training from scratch with $S_{KDE30}$ from \citep{schurholtHyperRepresentationsGenerativeModels2022}, \ourmethod combined with the $KDE30$ sampling method, and our \ourmethod subsampled. Each of the sampled populations is fine-tuned over 25 epochs.
}\label{tab:generative_cnn_zoos_in_domain}
\small
\setlength{\tabcolsep}{2.8pt}
\begin{tabularx}{1.0\linewidth}{clcccc}
\toprule
Ep.                  & \multicolumn{1}{c}{Method} & MNIST               & SVHN               & CIFAR-10           & STL                \\
\cmidrule(r){1-1} \cmidrule(lr){2-2} \cmidrule(lr){3-3} \cmidrule(lr){4-4} \cmidrule(lr){5-5} \cmidrule(l){6-6}
\multirow{5}{*}{0}     &  tr. fr. scratch                  & $\sim$10 /\%        & $\sim$10 /\%       & $\sim$10 /\%       & $\sim$10 /\%       \\
                       & $S_{KDE30}$                & 68.6$\pm$6.7           & 54.5$\pm$5.9          & \textit{n/a}                & \textit{n/a}                \\
                       & \ourmethod$_{KDE30}$                    & 84.8$\pm$0.8           & 70.7$\pm$1.4          & 56.3$\pm$0.5          & 39.2$\pm$0.8          \\
                       & \ourmethod$_{SUB}$                   & \textbf{86.7$\pm$0.8}  & \textbf{72.3$\pm$1.6} & \textbf{57.9$\pm$0.2} & \textbf{43.5$\pm$1.0} \\
                      & \ourmethod$_{GAUSS}$                 & 20.8$\pm$0.1                & 21.6$\pm$0.5                & 19.3$\pm$0.2               & 17.5$\pm$1.5               \\
\cmidrule(r){1-1} \cmidrule(lr){2-2} \cmidrule(lr){3-3} \cmidrule(lr){4-4} \cmidrule(lr){5-5} \cmidrule(l){6-6}
\multirow{5}{*}{1}     &  tr. fr. scratch                  & 20.6$\pm$1.6           & 19.4$\pm$0.6          & 37.2$\pm$1.4          & 21.3$\pm$1.6          \\
                       & $S_{KDE30}$                & 83.7$\pm$1.3           & 69.9$\pm$1.6          & \textit{n/a}                & \textit{n/a}                \\
                       & \ourmethod$_{KDE30}$                    & 85.5$\pm$0.8           & 71.3$\pm$1.4          & 58.2$\pm$0.2          & 43.5$\pm$0.7          \\
                       & \ourmethod$_{SUB}$                   & \textbf{87.5$\pm$0.6}  & \textbf{73.3$\pm$1.4} & \textbf{59.1$\pm$0.3} & \textbf{44.3$\pm$1.0} \\
                        & \ourmethod$_{GAUSS}$                 & 61.3$\pm$3.1                & 24.1$\pm$4.4                & 27.2$\pm$0.3               & 22.4$\pm$1.0               \\
\cmidrule(r){1-1} \cmidrule(lr){2-2} \cmidrule(lr){3-3} \cmidrule(lr){4-4} \cmidrule(lr){5-5} \cmidrule(l){6-6}
\multirow{5}{*}{5}     &  tr. fr. scratch                  & 36.7$\pm$5.2           & 23.5$\pm$4.7          & 48.5$\pm$1.0          & 31.6$\pm$4.2          \\
                       & $S_{KDE30}$                & \textbf{92.4$\pm$0.7} & 57.3$\pm$12.4         & \textit{n/a}                & \textit{n/a}                \\
                       & \ourmethod$_{KDE30}$                    & 87.5$\pm$0.7           & 72.2$\pm$1.2          & 58.8$\pm$0.4          & 45.2$\pm$0.6          \\
                       & \ourmethod$_{SUB}$                   & 89.0$\pm$0.4           & \textbf{73.6$\pm$1.5} & \textbf{59.6$\pm$0.3} & \textbf{45.3$\pm$0.9} \\
                       & \ourmethod$_{GAUSS}$                 & 83.4$\pm$0.8                & 35.6$\pm$8.9                & 43.3$\pm$0.3               & 34.2$\pm$0.7               \\
\cmidrule(r){1-1} \cmidrule(lr){2-2} \cmidrule(lr){3-3} \cmidrule(lr){4-4} \cmidrule(lr){5-5} \cmidrule(l){6-6}
\multirow{5}{*}{25}    &  tr. fr. scratch                  & 83.3$\pm$2.6           & 66.7$\pm$8.5          & 57.2$\pm$0.8          & 44.0$\pm$1.0          \\
                       & $S_{KDE30}$                & \textbf{93.0$\pm$0.7}           & 74.2$\pm$1.4          &       \textit{n/a}             &    \textit{n/a}                \\
                       & \ourmethod$_{KDE30}$                    & 92.0$\pm$0.3           & 74.7$\pm$0.8          & 60.2$\pm$0.6          & \textbf{48.4$\pm$0.5} \\
                       & \ourmethod$_{SUB}$                   & 92.3$\pm$0.4           & \textbf{75.1$\pm$1.0} & \textbf{61.2$\pm$0.1} & 48.0$\pm$0.4          \\
                       & \ourmethod$_{GAUSS}$                 & \textbf{94.2$\pm$0.4}       & 54.2$\pm$17.6               & 52.2$\pm$0.6               & 43.5$\pm$0.5               \\
\cmidrule(r){1-1} \cmidrule(lr){2-2} \cmidrule(lr){3-3} \cmidrule(lr){4-4} \cmidrule(lr){5-5} \cmidrule(l){6-6}
\multicolumn{1}{l}{50} &  tr. fr. scratch                  & 91.1$\pm$2.6           & 70.7$\pm$8.8          & 61.5$\pm$0.7          & 47.4$\pm$0.9      \\
\bottomrule
\end{tabularx}
\end{table} 

\textbf{Sampling High-Performing CNNs Zero-Shot.}
We begin with finetuning and transfer learning experiments on small CNNs from the modelzoo dataset to validate that the sequential decomposition for pretraining and sampling does not hurt performance.
The results of these experiments show dramatically improved performance zero-shot for fine-tuning and transfer learning over previous \textit{hyper-representations}; see Tables \ref{tab:generative_cnn_zoos_in_domain} and \ref{tab:generative_cnn_zoos_transfer}. 
At epoch 0, \ourmethod improves over previous \textit{hyper-representations} $S_{KDE30}$ by almost 20\%.
The effect becomes smaller during fine-tuning.
Nonetheless, \ourmethod consistently outperforms training from scratch with a higher epoch budget, often by several percentage points. 
This demonstrates on small CNNs that sequential pretraining and sampling of \ourmethod improves performance, particularly zero shot. 
This indicates the potential for scenarios with little labelled data.
\begin{table}[h]
\vspace{-2mm}
\captionof{table}{
Model generation on ResNet-18 model populations fine-tuned on the same task. We compare sampled models at different epochs with models trained from scratch.
}
\label{tab:generative_resnets_zoos_in_domain}
\small
\setlength{\tabcolsep}{3pt}
\begin{tabularx}{1.0\linewidth}{clccc}
\toprule
Epoch               & \multicolumn{1}{c}{Method} & CIFAR-10           & CIFAR-100          & Tiny-Imagenet      \\
\cmidrule(r){1-1} \cmidrule(lr){2-2} \cmidrule(lr){3-3} \cmidrule(lr){4-4} \cmidrule(l){5-5} 
\multirow{2}{*}{0}  & tr. fr. scratch                  & $\sim$10 /\%       & $\sim$1 /\%        & $\sim$0.5 /\%      \\
                    & \ourmethod$_{KDE30}$                    & 64.8$\pm$2.0          & 19.8$\pm$2.5          & 8.4$\pm$0.9           \\
                    & \ourmethod$_{SUB}$                   & 68.1$\pm$0.7          & 19.8$\pm$1.3          & 11.1$\pm$0.5          \\
                    & \ourmethod$_{BOOT}$                  & \textbf{68.6$\pm$1.2} & \textbf{20.4$\pm$1.3} & \textbf{11.7$\pm$0.5} \\
\cmidrule(r){1-1} \cmidrule(lr){2-2} \cmidrule(lr){3-3} \cmidrule(lr){4-4} \cmidrule(l){5-5} 
\multirow{2}{*}{1}  & tr. fr. scratch                  & 43.7$\pm$1.3          & 17.5$\pm$0.7          & 13.8$\pm$0.8          \\
                    & \ourmethod$_{KDE30}$                    & 82.4$\pm$0.9          & 59.0$\pm$1.3          & 46.7$\pm$0.8          \\
                    & \ourmethod$_{SUB}$                   & \textbf{83.6$\pm$1.5} & \textbf{60.8$\pm$0.8} & \textbf{47.4$\pm$1.0}          \\
                    & \ourmethod$_{BOOT}$                  & 82.8$\pm$1.4          & 60.2$\pm$0.5          & 47.2$\pm$0.8 \\
\cmidrule(r){1-1} \cmidrule(lr){2-2} \cmidrule(lr){3-3} \cmidrule(lr){4-4} \cmidrule(l){5-5} 
\multirow{2}{*}{5}  & tr. fr. scratch                  & 64.4$\pm$2.9          & 36.5$\pm$2.0          & 31.1$\pm$1.6          \\
                    & \ourmethod$_{KDE30}$                    & \textbf{85.9$\pm$0.6} & 56.2$\pm$1.7          & 45.6$\pm$1.4          \\
                    & \ourmethod$_{SUB}$                   & 85.4$\pm$1.3          &\textbf{ 56.7$\pm$1.6  }        & 45.7$\pm$0.8          \\
                    & \ourmethod$_{BOOT}$                  & 85.4$\pm$0.7          & 56.4$\pm$1.2          & \textbf{49.1$\pm$1.7} \\
\cmidrule(r){1-1} \cmidrule(lr){2-2} \cmidrule(lr){3-3} \cmidrule(lr){4-4} \cmidrule(l){5-5} 
\multirow{2}{*}{10} & tr. fr. scratch                  &  76.5$\pm$2.7           & 49.0$\pm$2.0        &  39.9$\pm$2.2       \\
                    & \ourmethod$_{KDE30}$                    & 91.4$\pm$0.1          & \textbf{72.9$\pm$0.2} & \textbf{64.2$\pm$0.3} \\
                    & \ourmethod$_{SUB}$                   & \textbf{91.6$\pm$0.2}          & \textbf{72.9$\pm$0.1} & 64.0$\pm$0.2          \\
                    & \ourmethod$_{BOOT}$                  & 91.6$\pm$0.2          & 72.8$\pm$0.1          & 64.1$\pm$0.2          \\
\cmidrule(r){1-1} \cmidrule(lr){2-2} \cmidrule(lr){3-3} \cmidrule(lr){4-4} \cmidrule(l){5-5} 
25                  & tr. fr. scratch                  & 85.5$\pm$1.5          & 56.5$\pm$2.0          & 43.3$\pm$1.9          \\
50                  & tr. fr. scratch                  & 92.14$\pm$0.2         & 70.7$\pm$0.4          & 57.3$\pm$0.6          \\
60                  & tr. fr. scratch                  &   \textit{n/a}                 & 74.2$\pm$0.3          & 63.9$\pm$0.5         
\\
\bottomrule
\end{tabularx}
\vspace{-4mm}
\end{table}

\textbf{\ourmethod Sequential Sampling Scales to ResNets.}
To evaluate how well sampling with \ourmethod scales to larger models, we continue with experiments on ResNet-18s. 
The results of these experiments Tables \ref{tab:generative_resnets_zoos_in_domain} and \ref{tab:generative_resnets_zoos_transfer} show that despite the long sequences, the sampled ResNet models perform well above random initialization. 
For example, sampled ResNet-18s achieve 68.1\% on CIFAR-10 without any fine-tuning (Table \ref{tab:generative_resnets_zoos_in_domain}). 
These models are at least three orders of magnitude larger than previous models used for \textit{hyper-representation} learning \citep{schurholtHyperRepresentationsGenerativeModels2022}, rendering it computationally infeasible for the approach presented in \citep{schurholtHyperRepresentationsGenerativeModels2022} to be evaluated against.
As before, the performance difference to random initialization becomes smaller during fine-tuning. 
Similar to our experiments on CNNs, sampled ResNet-18s achieve competitive performance or even outperform training from scratch with a considerably smaller computational budget.\footnote{The base population is trained with a one-cycle learning rate scheduler. To avoid any bias, we adopt the same scheduler but train for only 10 epochs, which affects direct comparability.}
Transferred to a new task, sampled models outperform training from scratch and match fine-tuning from prompt examples (Table \ref{tab:generative_resnets_zoos_transfer}). Interestingly, subsampling and bootstrapping appear to work well when there is a useful signal to start with, i.e., on easier tasks that are similar to the pretraining distribution. This suggests that the sampling distributions are not ideal, and may require a better fit, more samples, or iterative adjustment to fit new datasets zero-shot.
Nonetheless, even the relatively naive sampling methods can successfully sample competitive models, even at the scale of ResNet-sized architectures. This shows that our sequential sampling works even for long sequences of tokens.  

\textbf{Subsampling Improves Performance.}
Previous work requires high-quality prompt examples to target specific properties \citep{schurholtHyperRepresentationsGenerativeModels2022}. Our sampling methods drop these requirements and use prompt examples only to model a prior. We therefore compare \ourmethod with $S_{KDE30}$ from \citep{schurholtHyperRepresentationsGenerativeModels2022} to \ourmethod. Further, we compare the $KDE30$ sampling method with our subsampling approach on \ourmethod. 
On datasets where published results are available, using $KDE30$ with \ourmethod improves performance over previously published results with $S_{KDE30}$; see Table \ref{tab:generative_cnn_zoos_in_domain} for MNIST and SVHN results, e.g., epoch 0. We credit this to the better reconstruction quality of pre-training with \ourmethod.
Further, our sampling methods improve performance over $S_{KDE30}$. We compare \ourmethod + $S_{KDE30}$ with \ourmethod + subsampling and \ourmethod + bootstrapping, e.g., in Table \ref{tab:generative_resnets_zoos_in_domain} on CIFAR-10 at epoch 0 from 64.8\% to 68.1\%, or on Tiny Imagenet from 8.4\% to 11.1\%. 
Using bootstrapping to adjust $\mathcal{P}$ iteratively further improves the sampled models slightly. It even allows to replace prompt examples with a Gaussian prior $\mathcal{P}$. The results of \ourmethod$_{GAUSS}$ show high performance after fine-tuning, even the highest overall on MNIST. 
These results show that our sampling methods not only drop requirements for the prompt examples but even improve the performance of the sampled models.

\textbf{Few-Shot Model Sampling Transfers to New Tasks and Architectures.}
Lastly, we explore whether sampling models using \ourmethod generalizes beyond the original task and architecture with very few prompt examples. 
Such transfers are out of reach of previous \textit{hyper-representations}, which are bound to a fixed number of weights.
\ourmethod, on the other hand, represents models of different sizes or architectures simply as sequences of different lengths, which can vary between pretraining and sampling. 
Since we use the prompt examples only to roughly model the sampling distribution, we need only a few (1-5) prompt examples which are trained for only a few epochs (1-5). 
That way, sampling for new architectures and/or tasks can become very efficient.
We test that idea in three experiments: 
(i) \emph{changing the tasks} between pretraining and prompt-examples from CIFAR-100 to Tiny-Imagenet (Table \ref{tab:generative_resnets_zoos_fewshot_transfer}); 
(ii) \emph{changing the architecture} between pretraining and prompt-examples from ResNet-18 to ResNet-34 (Table \ref{tab:generative_resnets_zoos_fewshot_resnet34}); and 
(iii) \emph{changing both task and architecture} from ResNet-18 on CIFAR-100 to ResNet-34 on Tiny-Imagenet (Figure \ref{fig:sampling_new_arch_task} and Table \ref{tab:generative_resnets_zoos_fewshot_ti_resnet34}).

In all three experiments, using target prompt examples improves over random initialization as well as previous transfer experiments. This indicates that \ourmethod representations contain useful information even for new architectures or tasks. 
The sampled models outperform the prompt examples and training from scratch, considerably in earlier epochs, and preserve a performance advantage throughout fine-tining. 

\begin{wraptable}{r}{0.59\linewidth}
\vspace{-2mm}
\captionof{table}{
Sampling ResNet-18 models for Tiny-Imagenet. \ourmethod was pre-trained on CIFAR-100, 15 samples are drawn using subsampling, and 5 prompt examples are taken from the Tiny-Imagenet ResNet-18 zoo at epoch 25 with a mean accuracy of 43\%.  
}
\label{tab:generative_resnets_zoos_fewshot_transfer}
\small
\setlength{\tabcolsep}{3pt}
\begin{tabularx}{1.0\linewidth}{clc}
\toprule
\multicolumn{3}{c}{ResNet-18   CIFAR100 to TinyImagnet}                           \\
\midrule
Ep.               & Method                    & Acc TI \\
\cmidrule(r){1-1}  \cmidrule(l){2-2} \cmidrule(lr){3-3} 
\multirow{2}{*}{0}  & tr. fr. scratch           & 0.5$\pm$0.0           \\
                    & \ourmethod                & 0.6$\pm$0.0           \\
\cmidrule(r){1-1}  \cmidrule(l){2-2} \cmidrule(lr){3-3} 
\multirow{2}{*}{1}  & tr. fr. scratch           & 10.4$\pm$2.2          \\
                    & \ourmethod                & \textbf{39.4$\pm$1.5}  \\
\cmidrule(r){1-1}  \cmidrule(l){2-2} \cmidrule(lr){3-3} 
\multirow{2}{*}{2}  & tr. fr. scratch           & 28.5$\pm$0.9          \\
                    & \ourmethod                & \textbf{61.0$\pm$0.2}  \\
\cmidrule(r){1-1}  \cmidrule(l){2-2} \cmidrule(lr){3-3} 
2            & \ourmethod ensamble                          & 64.0     \\                          
\bottomrule
\end{tabularx}
\vspace{-2mm}
\end{wraptable} 
Sampling for a new task (Table~\ref{tab:generative_resnets_zoos_fewshot_transfer}), the sampled models outperform the prompt examples after just two epochs of fine-tuning, which indicates that transfer-learning using \ourmethod is an efficient alternative. 
Sampling from ResNet-18 to ResNet-34 for the same task (Table \ref{tab:generative_resnets_zoos_fewshot_resnet34}) shows likewise improved performance over training from scratch, which indicates that the learned representation generalizes to larger architectures as well.
Sampling for new tasks and different architecture (Figure \ref{fig:sampling_new_arch_task} and Table \ref{tab:generative_resnets_zoos_fewshot_ti_resnet34}) combines the previous experiments and confirms their results. Sampled models outperform training from scratch by a considerable margin. 
Figure \ref{fig:sampling_new_arch_task} indicates that with increasing distance from the pretraining architecture for \ourmethod, the performance gain of sampled models decreases, e.g., with increasing ResNet size.
Additionally, since sampling models using \ourmethod is cheap and lends itself to ensembling, we investigate the diversity of sampled models in Appendix \ref{sec:sampling_diversity}.
Taken together, the experiments show that \ourmethod learns representations that can generalize beyond the pretraining task and architecture, and can efficiently be sampled for both new tasks and architectures.

\subsection*{Limitations}
In this paper, we pretrain \ourmethod on homogeneous zoos with one architecture. This simplifies alignment for pre-training, but more importantly it simplifies evaluation for model generation. Since \ourmethod can train on varying architectures and model sizes, the model population requirement for pre-training is significantly relaxed. A sufficient number of models are available on public model hubs. 
Further, our sampling method requires access to prompt examples, to have an informed prior from which to sample. For small models, bootstrapping from a Gaussian finds the targeted distribution; see \ourmethod$_{GAUSS}$ in Table \ref{tab:generative_cnn_zoos_in_domain}. For large models with correspondingly long sequences, that approach is too expensive, which is why we rely on prompt examples.  
Lastly, in this paper, we perform experiments only on computer vision tasks. This is a choice to simplify the experiment setup.

%
%
%
%
\section{Related Work}

Representation learning in the space of NN weights has become a growing field recently. 
Several methods with different approaches to deal with weight spaces have been proposed to predict model properties such as accuracy \citep{unterthinerPredictingNeuralNetwork2020,eilertsenClassifyingClassifierDissecting2020, andreisSetbasedNeuralNetwork2023, zhangNeuralNetworksAre2023} or to learn the encoded concepts \citep{ashkenaziNeRNLearningNeural2022, deluigiDeepLearningImplicit2023}. 
Other work investigates the structure of trained weights on a fundamental level, using their eigen or singular value decompositions to identify training phases or predict properties \citep{martinTraditionalHeavyTailedSelf2019,MM20_SDM,martinPredictingTrendsQuality2021,martinImplicitSelfregularizationDeep2021, YTHx22_TR, mellerSingularValueRepresentation2023}.
Taking an optimization perspective, other work has investigated the uniqueness of the basis of trained NNs \citep{ainsworthGitReBasinMerging2022, brownPrivilegedConvergentBases2023}. Other work identifies subspaces of weights that are relevant, which motivates our work \citep{bentonLossSurfaceSimplexes2021, lucasMonotonicLinearInterpolation2021, wortsmanLearningNeuralNetwork2021, fortLargeScaleStructure2019}. The mode connectivity of trained models has been investigated to improve understanding of how to train models \citep{draxlerEssentiallyNoBarriers2018, nguyenConnectedSublevelSets2019, frankleLinearModeConnectivity2019}. 

A different line of work trains models to generate weights for target models, such as HyperNetworks \citep{haHyperNetworks2016, nguyenHyperVAEMinimumDescription2019, zhangGraphHyperNetworksNeural2019, knyazevParameterPredictionUnseen2021,knyazevCanWeScale2023b,kofinas2024graph}, with a recurrent backbone \citep{wangCompactOptimalDeep2023} as learned initialization \citep{dauphinMetaInitInitializingLearning2019} or for meta learning \citep{finnModelAgnosticMetaLearningFast2017,zhmoginovHyperTransformerModelGeneration2022,navaMetaLearningClassifierFree2022}.
While the last category uses data to get learning signals, another line of work learns representations of the weights directly. Hyper-Representations train an encoder-decoder architecture using reconstruction of the weights, with contrastive guidance, and has been proposed to predict model properties \citep{schurholtSelfSupervisedRepresentationLearning2021} or generate new models \citep{schurholtHyperRepresentationsPreTrainingTransfer2022, schurholtHyperRepresentationsGenerativeModels2022}. While previous work was limited to small models of fixed length, this paper proposes methods to decouple the representation learner size from the base model. Related approaches use convolutional auto-encoders \citep{berardiLearningSpaceDeep2022} or diffusion on the weights \citep{peeblesLearningLearnGenerative2022}.

%
%
\section{Conclusion}

In this work, we propose \ourmethod, a method to learn task-agnostic representations of Neural Network models. \ourmethod decouples model tokenization from \textit{hyper-representation} learning and can scale to much larger neural network models and generalize to models of different architectures. 
We analyze \ourmethod embeddings and find they reveal model quality metrics. Empirical evaluations show that 
i) \ourmethod embeddings contain information on model quality both globally and on a layer level,
ii) \ourmethod embeddings are predictive of model performance, and 
iii) sampling models with \ourmethod achieves higher performance and generalizes to larger models and new architectures. Further, we propose sampling methods that reduce quality and quantity requirements for prompt examples and allow targeting new model distributions.\looseness-1

\newpage
\subsection*{Acknowledgements}
KS and DB would like to acknowledge the Google Research Scholar Award and the Swiss National Science Foundation for partial funding of this work. We are thankful to Erik Vee at Google Research for the insightful discussions, Léo Meynent and Joelle Hanna for editorial support, and Harald Rotter, Kurt Städler, and Martin Eigenmann for the support with the computational infrastructure which made this project possible.
MWM would like to acknowledge the NSF, DOE, and IARPA for partial support of this work.

\subsection*{Impact Statement}
This paper introduces a novel weight representation learning method designed to enhance the performance and scalability of machine learning models across various applications. As a fundamental approach, it serves as a foundation for future advancements in the field of machine learning. \ourmethod holds potential for use in both academic research and industry applications and thus inherits all their benefits but also risks for adverse applications of machine learning. Its versatility and scalability make \ourmethod a valuable tool that may also offer insights into model interpretability.

\bibliography{./bib_auto.bib,./bib_manual.bib}

\begin{thebibliography}{53}
\providecommand{\natexlab}[1]{#1}
\providecommand{\url}[1]{\texttt{#1}}
\expandafter\ifx\csname urlstyle\endcsname\relax
  \providecommand{\doi}[1]{doi: #1}\else
  \providecommand{\doi}{doi: \begingroup \urlstyle{rm}\Url}\fi

\bibitem[Ainsworth et~al.(2022)Ainsworth, Hayase, and Srinivasa]{ainsworthGitReBasinMerging2022}
Ainsworth, S.~K., Hayase, J., and Srinivasa, S.
\newblock Git {{Re-Basin}}: {{Merging Models}} modulo {{Permutation Symmetries}}, September 2022.

\bibitem[Andreis et~al.(2023)Andreis, Bedionita, and Hwang]{andreisSetbasedNeuralNetwork2023}
Andreis, B., Bedionita, S., and Hwang, S.~J.
\newblock Set-based {{Neural Network Encoding}}, May 2023.

\bibitem[Ashkenazi et~al.(2022)Ashkenazi, Rimon, Vainshtein, Levi, Richardson, Mintz, and Treister]{ashkenaziNeRNLearningNeural2022}
Ashkenazi, M., Rimon, Z., Vainshtein, R., Levi, S., Richardson, E., Mintz, P., and Treister, E.
\newblock {{NeRN}} -- {{Learning Neural Representations}} for {{Neural Networks}}, December 2022.

\bibitem[Benton et~al.(2021)Benton, Maddox, Lotfi, and Wilson]{bentonLossSurfaceSimplexes2021}
Benton, G.~W., Maddox, W.~J., Lotfi, S., and Wilson, A.~G.
\newblock Loss {{Surface Simplexes}} for {{Mode Connecting Volumes}} and {{Fast Ensembling}}.
\newblock In \emph{{{PMLR}}}, 2021.

\bibitem[Berardi et~al.(2022)Berardi, De~Luigi, Salti, and Di~Stefano]{berardiLearningSpaceDeep2022}
Berardi, G., De~Luigi, L., Salti, S., and Di~Stefano, L.
\newblock Learning the {{Space}} of {{Deep Models}}, June 2022.

\bibitem[Bishop(2006)]{bishopPatternRecognitionMachine2006}
Bishop, C.~M.
\newblock \emph{Pattern Recognition and Machine Learning}.
\newblock {springer}, 2006.

\bibitem[Brown et~al.(2023)Brown, Vyas, and Bansal]{brownPrivilegedConvergentBases2023}
Brown, D., Vyas, N., and Bansal, Y.
\newblock On {{Privileged}} and {{Convergent Bases}} in {{Neural Network Representations}}, July 2023.

\bibitem[Corneanu et~al.(2020)Corneanu, Escalera, and Martinez]{corneanuComputingTestingError2020}
Corneanu, C.~A., Escalera, S., and Martinez, A.~M.
\newblock Computing the {{Testing Error Without}} a {{Testing Set}}.
\newblock In \emph{2020 {{IEEE}}/{{CVF Conference}} on {{Computer Vision}} and {{Pattern Recognition}} ({{CVPR}})}. {IEEE}, June 2020.

\bibitem[Dao et~al.(2022)Dao, Fu, Ermon, Rudra, and R{\'e}]{daoFlashAttentionFastMemoryEfficient2022}
Dao, T., Fu, D.~Y., Ermon, S., Rudra, A., and R{\'e}, C.
\newblock {{FlashAttention}}: {{Fast}} and {{Memory-Efficient Exact Attention}} with {{IO-Awareness}}, June 2022.

\bibitem[Dauphin \& Schoenholz(2019)Dauphin and Schoenholz]{dauphinMetaInitInitializingLearning2019}
Dauphin, Y.~N. and Schoenholz, S.
\newblock {{MetaInit}}: {{Initializing}} learning by learning to initialize.
\newblock In \emph{Neural {{Information Processing Systems}}}, 2019.

\bibitem[De~Luigi et~al.(2023)De~Luigi, Cardace, Spezialetti, Ramirez, Salti, and Di~Stefano]{deluigiDeepLearningImplicit2023}
De~Luigi, L., Cardace, A., Spezialetti, R., Ramirez, P.~Z., Salti, S., and Di~Stefano, L.
\newblock Deep {{Learning}} on {{Implicit Neural Representations}} of {{Shapes}}, February 2023.

\bibitem[Deutsch(2018)]{deutschGeneratingNeuralNetworks2018}
Deutsch, L.
\newblock Generating {{Neural Networks}} with {{Neural Networks}}.
\newblock April 2018.

\bibitem[Draxler et~al.(2018)Draxler, Veschgini, Salmhofer, and Hamprecht]{draxlerEssentiallyNoBarriers2018}
Draxler, F., Veschgini, K., Salmhofer, M., and Hamprecht, F.
\newblock Essentially {{No Barriers}} in {{Neural Network Energy Landscape}}.
\newblock In \emph{International {{Conference}} on {{Machine Learning}}}, March 2018.

\bibitem[Eilertsen et~al.(2020)Eilertsen, J{\"o}nsson, Ropinski, Unger, and Ynnerman]{eilertsenClassifyingClassifierDissecting2020}
Eilertsen, G., J{\"o}nsson, D., Ropinski, T., Unger, J., and Ynnerman, A.
\newblock Classifying the classifier: Dissecting the weight space of neural networks.
\newblock February 2020.

\bibitem[Finn et~al.(2017)Finn, Abbeel, and Levine]{finnModelAgnosticMetaLearningFast2017}
Finn, C., Abbeel, P., and Levine, S.
\newblock Model-{{Agnostic Meta-Learning}} for {{Fast Adaptation}} of {{Deep Networks}}.
\newblock In \emph{Proceedings of the 34th {{International Conference}} on {{Machine Learning}}}. {PMLR}, July 2017.

\bibitem[Fort \& Jastrzebski(2019)Fort and Jastrzebski]{fortLargeScaleStructure2019}
Fort, S. and Jastrzebski, S.
\newblock Large {{Scale Structure}} of {{Neural Network Loss Landscapes}}.
\newblock June 2019.

\bibitem[Frankle et~al.(2019)Frankle, Dziugaite, Roy, and Carbin]{frankleLinearModeConnectivity2019}
Frankle, J., Dziugaite, G., Roy, D.~M., and Carbin, M.
\newblock Linear {{Mode Connectivity}} and the {{Lottery Ticket Hypothesis}}.
\newblock December 2019.

\bibitem[Ha et~al.(2016)Ha, Dai, and Le]{haHyperNetworks2016}
Ha, D., Dai, A., and Le, Q.~V.
\newblock {{HyperNetworks}}, 2016.

\bibitem[Jiang et~al.(2019)Jiang, Krishnan, Mobahi, and Bengio]{jiangPredictingGeneralizationGap2019}
Jiang, Y., Krishnan, D., Mobahi, H., and Bengio, S.
\newblock Predicting the {{Generalization Gap}} in {{Deep Networks}} with {{Margin Distributions}}.
\newblock June 2019.

\bibitem[Knyazev et~al.(2021)Knyazev, Drozdzal, Taylor, and {Romero-Soriano}]{knyazevParameterPredictionUnseen2021}
Knyazev, B., Drozdzal, M., Taylor, G.~W., and {Romero-Soriano}, A.
\newblock Parameter {{Prediction}} for {{Unseen Deep Architectures}}.
\newblock In \emph{Conference on {{Neural Information Processing Systems}} ({{NeurIPS}})}, 2021.

\bibitem[Knyazev et~al.(2023)Knyazev, Hwang, and {Lacoste-Julien}]{knyazevCanWeScale2023b}
Knyazev, B., Hwang, D., and {Lacoste-Julien}, S.
\newblock Can {{We Scale Transformers}} to {{Predict Parameters}} of {{Diverse ImageNet Models}}?
\newblock In \emph{{{arXiv}}.Org}, March 2023.

\bibitem[Kofinas et~al.(2024)Kofinas, Knyazev, Zhang, Chen, Burghouts, Gavves, Snoek, and Zhang]{kofinas2024graph}
Kofinas, M., Knyazev, B., Zhang, Y., Chen, Y., Burghouts, G.~J., Gavves, E., Snoek, C. G.~M., and Zhang, D.~W.
\newblock Graph neural networks for learning equivariant representations of neural networks.
\newblock In \emph{International {{Conference}} on {{Learning Representations}} ({{ICLR}})}, 2024.

\bibitem[Kornblith et~al.(2019)Kornblith, Norouzi, Lee, and Hinton]{kornblithSimilarityNeuralNetwork2019}
Kornblith, S., Norouzi, M., Lee, H., and Hinton, G.
\newblock Similarity of {{Neural Network Representations Revisited}}.
\newblock May 2019.

\bibitem[Leclerc et~al.(2023)Leclerc, Ilyas, Engstrom, Park, Salman, and Madry]{leclercFFCVAcceleratingTraining2023}
Leclerc, G., Ilyas, A., Engstrom, L., Park, S.~M., Salman, H., and Madry, A.
\newblock {{FFCV}}: {{Accelerating Training}} by {{Removing Data Bottleneck}}.
\newblock In \emph{Computer {{Vision}} and {{Pattern Recognition}} ({{CVPR}})}, 2023.

\bibitem[Liaw et~al.(2018)Liaw, Liang, Nishihara, Moritz, Gonzalez, and Stoica]{liawTuneResearchPlatform2018}
Liaw, R., Liang, E., Nishihara, R., Moritz, P., Gonzalez, J.~E., and Stoica, I.
\newblock Tune: {{A Research Platform}} for {{Distributed Model Selection}} and {{Training}}.
\newblock July 2018.

\bibitem[Lucas et~al.(2021)Lucas, Bae, Zhang, Fort, Zemel, and Grosse]{lucasMonotonicLinearInterpolation2021}
Lucas, J.~R., Bae, J., Zhang, M.~R., Fort, S., Zemel, R., and Grosse, R.~B.
\newblock On {{Monotonic Linear Interpolation}} of {{Neural Network Parameters}}.
\newblock In \emph{International {{Conference}} on {{Machine Learning}}}. {PMLR}, July 2021.

\bibitem[Martin \& Mahoney(2019{\natexlab{a}})Martin and Mahoney]{martinRethinkingGeneralizationRequires2019}
Martin, C.~H. and Mahoney, M.~W.
\newblock Rethinking generalization requires revisiting old ideas: Statistical mechanics approaches and complex learning behavior, February 2019{\natexlab{a}}.

\bibitem[Martin \& Mahoney(2019{\natexlab{b}})Martin and Mahoney]{martinTraditionalHeavyTailedSelf2019}
Martin, C.~H. and Mahoney, M.~W.
\newblock Traditional and {{Heavy-Tailed Self Regularization}} in {{Neural Network Models}}.
\newblock January 2019{\natexlab{b}}.

\bibitem[Martin \& Mahoney(2020)Martin and Mahoney]{MM20_SDM}
Martin, C.~H. and Mahoney, M.~W.
\newblock Heavy-tailed {U}niversality predicts trends in test accuracies for very large pre-trained deep neural networks.
\newblock In \emph{Proceedings of the 20th SIAM International Conference on Data Mining}, 2020.

\bibitem[Martin \& Mahoney(2021)Martin and Mahoney]{martinImplicitSelfregularizationDeep2021}
Martin, C.~H. and Mahoney, M.~W.
\newblock Implicit self-regularization in deep neural networks: Evidence from random matrix theory and implications for learning.
\newblock \emph{The Journal of Machine Learning Research}, 22\penalty0 (1), January 2021.

\bibitem[Martin et~al.(2021)Martin, Peng, and Mahoney]{martinPredictingTrendsQuality2021}
Martin, C.~H., Peng, T.~S., and Mahoney, M.~W.
\newblock Predicting trends in the quality of state-of-the-art neural networks without access to training or testing data.
\newblock \emph{Nature Communications}, 12\penalty0 (1), July 2021.

\bibitem[Meller \& Berkouk(2023)Meller and Berkouk]{mellerSingularValueRepresentation2023}
Meller, D. and Berkouk, N.
\newblock Singular {{Value Representation}}: {{A New Graph Perspective On Neural Networks}}, February 2023.

\bibitem[Nava et~al.(2022)Nava, Kobayashi, Yin, Katzschmann, and Grewe]{navaMetaLearningClassifierFree2022}
Nava, E., Kobayashi, S., Yin, Y., Katzschmann, R.~K., and Grewe, B.~F.
\newblock Meta-{{Learning}} via {{Classifier}}(-free) {{Diffusion Guidance}}.
\newblock October 2022.

\bibitem[Nguyen et~al.(2019)Nguyen, Tran, Gupta, Rana, and Dam]{nguyenHyperVAEMinimumDescription2019}
Nguyen, P., Tran, T., Gupta, S., Rana, S., and Dam, H.-C.
\newblock {{HyperVAE}}: {{A Minimum Description Length Variational Hyper-Encoding Network}}.
\newblock 2019.

\bibitem[Nguyen(2019)]{nguyenConnectedSublevelSets2019}
Nguyen, Q.~N.
\newblock On {{Connected Sublevel Sets}} in {{Deep Learning}}.
\newblock In \emph{International {{Conference}} on {{Machine Learning}}}, January 2019.

\bibitem[Paszke et~al.(2019)Paszke, Gross, Massa, Lerer, Bradbury, Chanan, Killeen, Lin, Gimelshein, Antiga, Desmaison, Kopf, Yang, DeVito, Raison, Tejani, Chilamkurthy, Steiner, Fang, Bai, and Chintala]{paszkePyTorchImperativeStyle2019}
Paszke, A., Gross, S., Massa, F., Lerer, A., Bradbury, J., Chanan, G., Killeen, T., Lin, Z., Gimelshein, N., Antiga, L., Desmaison, A., Kopf, A., Yang, E., DeVito, Z., Raison, M., Tejani, A., Chilamkurthy, S., Steiner, B., Fang, L., Bai, J., and Chintala, S.
\newblock {{PyTorch}}: {{An Imperative Style}}, {{High-Performance Deep Learning Library}}.
\newblock In \emph{Advances in {{Neural Information Processing Systems}} 32}, 2019.

\bibitem[Peebles et~al.(2022)Peebles, Radosavovic, Brooks, Efros, and Malik]{peeblesLearningLearnGenerative2022}
Peebles, W., Radosavovic, I., Brooks, T., Efros, A.~A., and Malik, J.
\newblock Learning to {{Learn}} with {{Generative Models}} of {{Neural Network Checkpoints}}, September 2022.

\bibitem[Ratzlaff \& Fuxin(2019)Ratzlaff and Fuxin]{ratzlaffHyperGANGenerativeModel2019}
Ratzlaff, N. and Fuxin, L.
\newblock {{HyperGAN}}: {{A Generative Model}} for {{Diverse}}, {{Performant Neural Networks}}.
\newblock In \emph{Proceedings of the 36th {{International Conference}} on {{Machine Learning}}}. {PMLR}, May 2019.

\bibitem[Sch{\"u}rholt et~al.(2021)Sch{\"u}rholt, Kostadinov, and Borth]{schurholtSelfSupervisedRepresentationLearning2021}
Sch{\"u}rholt, K., Kostadinov, D., and Borth, D.
\newblock Self-{{Supervised Representation Learning}} on {{Neural Network Weights}} for {{Model Characteristic Prediction}}.
\newblock In \emph{Conference on {{Neural Information Processing Systems}} ({{NeurIPS}})}, volume~35, 2021.

\bibitem[Sch{\"u}rholt et~al.(2022{\natexlab{a}})Sch{\"u}rholt, Knyazev, {Gir{\'o}-i-Nieto}, and Borth]{schurholtHyperRepresentationsGenerativeModels2022}
Sch{\"u}rholt, K., Knyazev, B., {Gir{\'o}-i-Nieto}, X., and Borth, D.
\newblock Hyper-{{Representations}} as {{Generative Models}}: {{Sampling Unseen Neural Network Weights}}.
\newblock In \emph{Thirty-Sixth {{Conference}} on {{Neural Information Processing Systems}} ({{NeurIPS}})}, September 2022{\natexlab{a}}.

\bibitem[Sch{\"u}rholt et~al.(2022{\natexlab{b}})Sch{\"u}rholt, Knyazev, {Gir{\'o}-i-Nieto}, and Borth]{schurholtHyperRepresentationsPreTrainingTransfer2022}
Sch{\"u}rholt, K., Knyazev, B., {Gir{\'o}-i-Nieto}, X., and Borth, D.
\newblock Hyper-{{Representations}} for {{Pre-Training}} and {{Transfer Learning}}.
\newblock In \emph{First {{Workshop}} of {{Pre-training}}: {{Perspectives}}, {{Pitfalls}}, and {{Paths Forward}} at {{ICML}} 2022}, 2022{\natexlab{b}}.

\bibitem[Sch{\"u}rholt et~al.(2022{\natexlab{c}})Sch{\"u}rholt, Taskiran, Knyazev, {Gir{\'o}-i-Nieto}, and Borth]{schurholtModelZoosDataset2022}
Sch{\"u}rholt, K., Taskiran, D., Knyazev, B., {Gir{\'o}-i-Nieto}, X., and Borth, D.
\newblock Model {{Zoos}}: {{A Dataset}} of {{Diverse Populations}} of {{Neural Network Models}}.
\newblock In \emph{Thirty-Sixth {{Conference}} on {{Neural Information Processing Systems}} ({{NeurIPS}}) {{Datasets}} and {{Benchmarks Track}}}, September 2022{\natexlab{c}}.

\bibitem[S{\'e}mery(2024)]{semeryOsmrImgclsmob2024}
S{\'e}mery, O.
\newblock Osmr/imgclsmob, January 2024.

\bibitem[Smith \& Topin(2018)Smith and Topin]{smithSuperConvergenceVeryFast2018}
Smith, L.~N. and Topin, N.
\newblock Super-{{Convergence}}: {{Very Fast Training}} of {{Neural Networks Using Large Learning Rates}}, May 2018.

\bibitem[Unterthiner et~al.(2020)Unterthiner, Keysers, Gelly, Bousquet, and Tolstikhin]{unterthinerPredictingNeuralNetwork2020}
Unterthiner, T., Keysers, D., Gelly, S., Bousquet, O., and Tolstikhin, I.
\newblock Predicting {{Neural Network Accuracy}} from {{Weights}}.
\newblock February 2020.

\bibitem[Vaswani et~al.(2021)Vaswani, Ramachandran, Srinivas, Parmar, Hechtman, and Shlens]{vaswaniScalingLocalSelfAttention2021}
Vaswani, A., Ramachandran, P., Srinivas, A., Parmar, N., Hechtman, B., and Shlens, J.
\newblock Scaling {{Local Self-Attention}} for {{Parameter Efficient Visual Backbones}}.
\newblock In \emph{Proceedings of the {{IEEE}}/{{CVF Conference}} on {{Computer Vision}} and {{Pattern Recognition}}}, 2021.

\bibitem[Wang et~al.(2023)Wang, Chen, Yu, Cheung, and LeCun]{wangCompactOptimalDeep2023}
Wang, J., Chen, Y., Yu, S.~X., Cheung, B., and LeCun, Y.
\newblock Compact and {{Optimal Deep Learning}} with {{Recurrent Parameter Generators}}.
\newblock In \emph{2023 {{IEEE}}/{{CVF Winter Conference}} on {{Applications}} of {{Computer Vision}} ({{WACV}})}. {IEEE}, January 2023.

\bibitem[Wortsman et~al.(2021)Wortsman, Horton, Guestrin, Farhadi, and Rastegari]{wortsmanLearningNeuralNetwork2021}
Wortsman, M., Horton, M.~C., Guestrin, C., Farhadi, A., and Rastegari, M.
\newblock Learning {{Neural Network Subspaces}}.
\newblock In \emph{International {{Conference}} on {{Machine Learning}}}. {PMLR}, July 2021.

\bibitem[Yak et~al.(2019)Yak, Gonzalvo, and Mazzawi]{yakTaskArchitectureIndependentGeneralization2019}
Yak, S., Gonzalvo, J., and Mazzawi, H.
\newblock Towards {{Task}} and {{Architecture-Independent Generalization Gap Predictors}}.
\newblock June 2019.

\bibitem[Yang et~al.(2022)Yang, Theisen, Hodgkinson, Gonzalez, Ramchandran, Martin, and Mahoney]{YTHx22_TR}
Yang, Y., Theisen, R., Hodgkinson, L., Gonzalez, J.~E., Ramchandran, K., Martin, C.~H., and Mahoney, M.~W.
\newblock Evaluating natural language processing models with generalization metrics that do not need access to any training or testing data.
\newblock Technical Report Preprint: arXiv:2202.02842, 2022.

\bibitem[Zhang et~al.(2019)Zhang, Ren, and Urtasun]{zhangGraphHyperNetworksNeural2019}
Zhang, C., Ren, M., and Urtasun, R.
\newblock Graph {{HyperNetworks}} for {{Neural Architecture Search}}.
\newblock In \emph{International {{Conference}} on {{Learning Representations}} ({{ICLR}})}, 2019.

\bibitem[Zhang et~al.(2023)Zhang, Kofinas, Zhang, Chen, Burghouts, and Snoek]{zhangNeuralNetworksAre2023}
Zhang, D.~W., Kofinas, M., Zhang, Y., Chen, Y., Burghouts, G.~J., and Snoek, C. G.~M.
\newblock Neural {{Networks Are Graphs}}!{{Graph Neural Networks}} for {{Equivariant Processing}} of {{Neural Networks}}.
\newblock July 2023.

\bibitem[Zhmoginov et~al.(2022)Zhmoginov, Sandler, and Vladymyrov]{zhmoginovHyperTransformerModelGeneration2022}
Zhmoginov, A., Sandler, M., and Vladymyrov, M.
\newblock {{HyperTransformer}}: {{Model Generation}} for {{Supervised}} and {{Semi-Supervised Few-Shot Learning}}.
\newblock In \emph{International {{Conference}} on {{Machine Learning}} ({{ICML}})}, January 2022.

\end{thebibliography}
\bibliographystyle{icml2024}

\newpage
\appendix
\onecolumn

\section{Ablation Studies}
\label{sec:ablation}
In this section, we perform ablation studies to assess the effectiveness of the methods proposed above: 
model alignment to simplify the learning task; 
inference window size to improve inference quality; 
haloing and batch-norm conditioning to increase sample quality.
\paragraph{Impact of Model Alignment.}
\begin{wraptable}{r}{0.38\linewidth}
\vspace{-8mm}
\captionof{table}{
Impact of alignment ablation and permutation on reconstruction loss.
}
\label{tab:ablation_alignment}
\small
\setlength{\tabcolsep}{6pt}
\begin{tabularx}{1.0\linewidth}{ccccc}
\toprule
\multicolumn{3}{c}{Sample Permutations} & \multicolumn{2}{c}{$\mathcal{L}_{rec}$} \\
\cmidrule(r){1-3} \cmidrule(l){4-5}
Aligned    & View 1       & View 2       & Train              & Test               \\
\cmidrule(r){1-3} \cmidrule(l){4-5}
No         & Perm.     & Perm.    & 0.304              & 0.167              \\
Yes        & Perm.     & Perm.    & 0.148              & 0.082              \\
Yes        & Align    & Perm.    & 0.107              & 0.082              \\
Yes        & Align    & Align   & 0.072              & 0.082              \\ 
\bottomrule
\end{tabularx}
\vspace{-4mm}
\end{wraptable} 
Model alignment intuitively reduces training complexity by mapping all models to the same subspace. To evaluate its impact, we conduct training experiments with the same configuration on datasets with and without aligned models. In the dataset with aligned models, we use either the aligned form or 5 random permutations for the two views for both reconstruction and contrastive learning.
As shown in Table \ref{tab:ablation_alignment}, the results show two effects. 
First, alignment through git re-basin simplifies the learning task and contributes to improved generalization, both training and test losses are reduced by more than 50\%. 
Second, anchoring at least one of the views to the aligned form does further reduce the training loss, but does not improve generalization.

\begin{wrapfigure}{r}{0.35\linewidth}
\vspace{-4mm}
\centering
\includegraphics[trim=4mm 0mm 1.5mm 1.5mm, clip, width=1.0\linewidth]{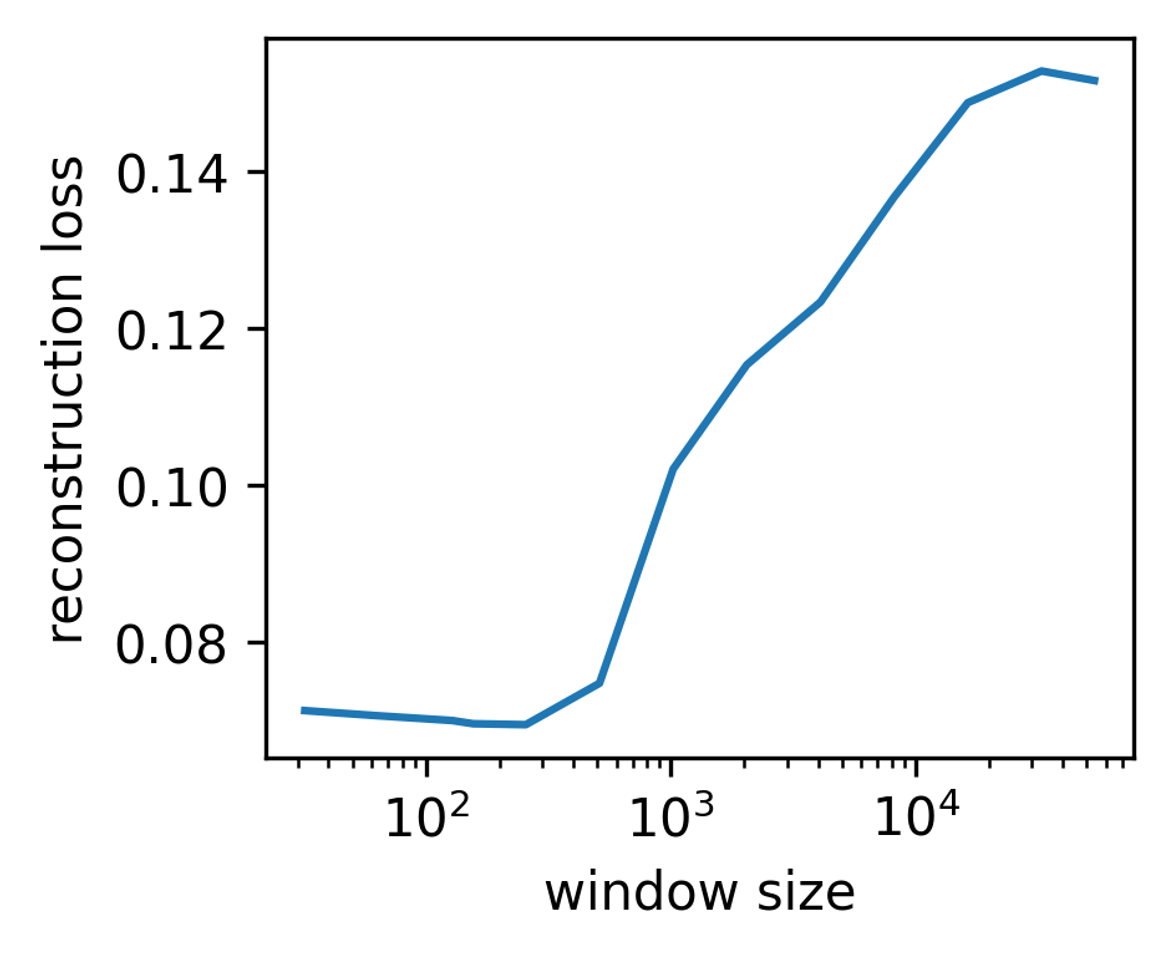}
\vspace{-8mm}
\captionof{figure}{\ourmethod reconstruction loss over number of tokens within a window. The loss is lowest around the training windowsize of 256 tokens, longer sequences up to the full model sequence length of 50k tokens cause interference and double the reconstruction error.
}
\vspace{-10mm}
\label{fig:windowsize_ablation}    
\end{wrapfigure}
\paragraph{Window Size Ablation.}
The sequential decomposition of \ourmethod allows one to pretrain not on the full model sequence, but on subsequences. The choice of the length of the subsequence, the window size, is a critical parameter that balances computational load and context. We used a window of 256 for pretraining for most of our experiments. 

Here, we study the influence of the window size on reconstruction error, exploring values ranging from 32 to 2048. Our experiments did not reveal substantial impact of smaller windows on pretraining loss or sampling performance. This seems to suggest that a window size as large as 2048 may still be insufficient on ResNets to capture enough context. Alternatively, it may suggest that the underlying assumption that context matters may not entirely hold up.

However, we did observe an important impact on the relationship between training and inference window sizes. During inference, memory load is significantly lower. Inference allows much larger window sizes, up to the entire length of the ResNet sequence. However, departing from the training window size appears to introduce interference, which affects the reconstruction error (Figure \ref{fig:windowsize_ablation}). 

\paragraph{Halo and batch-norm conditioning.}
\begin{wraptable}{r}{0.35\linewidth}
\vspace{-10mm}
\captionof{table}{
Ablation of batch-norm conditioning and haloing.
}\label{tab:ablation_halo_bn_cond}
\small
\setlength{\tabcolsep}{3pt}
\begin{tabularx}{1.0\linewidth}{clc}
\toprule
Ep.              & Method        & CIFAR-10           \\
\cmidrule(r){1-1} \cmidrule(lr){2-2} \cmidrule(l){3-3}
\multirow{5}{*}{0} & rand init     & $\sim$10 /\%       \\
                   & naïve         & 10$\pm$0.0            \\
\cmidrule(lr){2-2} \cmidrule(l){3-3}
                   
                   & Haloed        & 14.5$\pm$6.3          \\
                   & BN-cond       & 60.8$\pm$2.2          \\
                   & Haloed+BN-cond & \textbf{64.8$\pm$2.1} \\
\cmidrule(r){1-1} \cmidrule(lr){2-2} \cmidrule(l){3-3}
\multirow{5}{*}{5} & rand init     & 64.4$\pm$2.9          \\
                   & naïve         & 90.8$\pm$0.2          \\
\cmidrule(lr){2-2} \cmidrule(l){3-3}
                   & Haloed        & 90.9$\pm$0.1          \\
                   & BN-cond       & 90.7$\pm$0.2          \\
                   & Haloed+BN-cond & 90.9$\pm$0.2         
    \\ 
\bottomrule
\end{tabularx}
\vspace{-4mm}
\end{wraptable} 
Haloing and batch-norm conditioning aim at reducing noise in model sampling; see Section \ref{sec:seq_hyper_reps}. 
To assess their impact on sampling performance, we conduct an in-domain experiment using \ourmethod trained on CIFAR-10 ResNet-18s, using prompt examples from the train set and fine-tuning on CIFAR-10. We compare with \emph{naïve} sampling without haloing and batch-norm conditioning.
The results in Table \ref{tab:ablation_halo_bn_cond} show the significant improvements achieved by both haloing and batch-norm conditioning. From random guessing of naïve sampling, combining both improves to around 65\%. Since both methods aim at reducing noise for zero-shot sampling, their effect is largest then and diminishes somewhat during finetuning. Both methods not only improve zero-shot sampling per se but make the sampled models provide enough signal to facilitate subsampling or bootstrapping strategies.\looseness-1

\vspace{32pt}
\section{\ourmethod Architecture Details}

\begin{wraptable}{r}{0.45\linewidth}
\vspace{-16mm}
\captionof{table}{
Architecture Details for \ourmethod
}
\label{tab:architecture_details}
\small
\setlength{\tabcolsep}{12pt}
\begin{tabularx}{1.0\linewidth}{lcc}
\toprule
\multicolumn{1}{c}{Hyper-Parameter} & CNNs     & ResNet-18 \\ \midrule
tokensize                           & 289      & 288       \\
sequence lenght                     & $\sim$50 & $\sim$50k \\
window size                         & 32       & 256, 512  \\
d\_model                            & 1024     & 2048      \\
latent\_dim                         & 128      & 128       \\
transformer layers                  & 4        & 8         \\
transformer heads                   & 4,8      & 4,8       \\
\bottomrule
\end{tabularx}
\vspace{-8mm}
\end{wraptable} 
In Table \ref{tab:architecture_details}, we provide additional information on the training hyper-parameters for \ourmethod on populations of small CNNs as well as ResNet18s. These values are the stable mean across all experiments, exact values can vary from population to population. Full experiment configurations are documented in the code.

\vspace{16pt}

\section{\ourmethod Embedding Analysis - Additional Results}
\begin{figure}[h!]
\centering
\includegraphics[trim=4mm 0mm 1.5mm 1.5mm, width=0.6\linewidth]{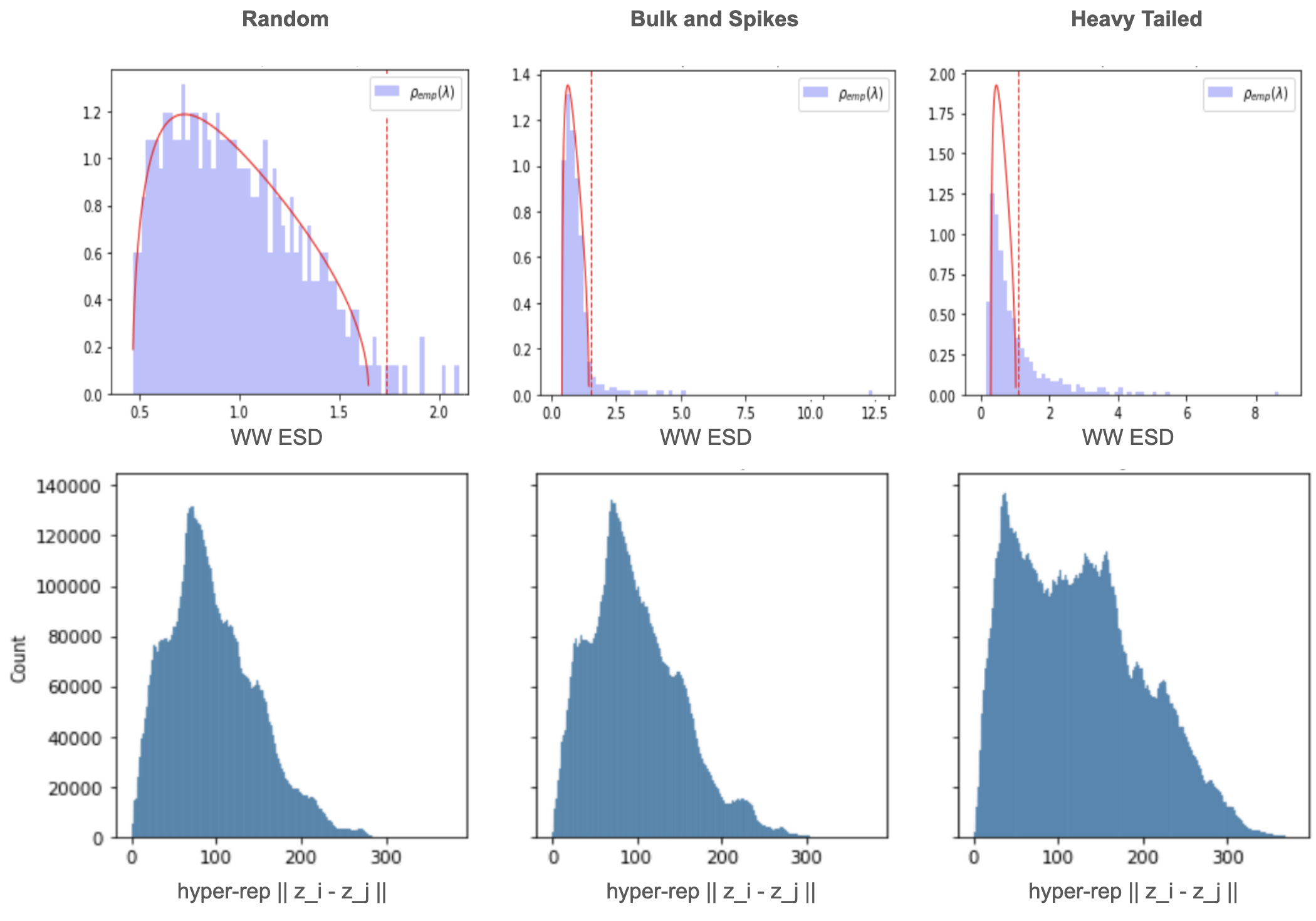}
\captionof{figure}{Comparison between WeightWatcher features (top) and \ourmethod (bottom). \citet{martinTraditionalHeavyTailedSelf2019} identify different phases in the eigenvalue spectrum of trained weight matrices. We replicate the experiment setup and find ESDs similar to \textit{random} (top left), \textit{bulk and spikes} (top middle) and \textit{heavy-tailed} (top right). We compare these against pairwise distances of \ourmethod embeddings of the same layer. While the distributions have a different shape, it appears to become more heavy-tailed going from \textit{random} to \textit{heavy tailed}.
}
\vspace{-2mm}
\label{fig:ww_comparison_phases}    
\end{figure}

This section contains additional results on \ourmethod embedding analysis, in comparison with previous weight matrix analysis. 
In Figure \ref{fig:ww_comparison_phases}, we compare the eigenvalue distribution for different models with \ourmethod embeddings. Replicating the experiment setup from \citep{martinTraditionalHeavyTailedSelf2019,martinImplicitSelfregularizationDeep2021}, we train MiniAlexNet models on CIFAR-10 varying only the batch size. With a smaller batch size and longer training duration, the eigenvalue distribution transitions from random with very few spikes, over a bulk with many spikes, to heavy-tailed. The embeddings of \ourmethod appear to also become more heavy-tailed, but do not seem to pick up on the change from few to many spikes. 
The results are suggestive, pointing to obvious follow-up work.

\begin{figure}[h!]
\centering
\includegraphics[trim=4mm 0mm 1.5mm 1.5mm, width=0.6\linewidth]{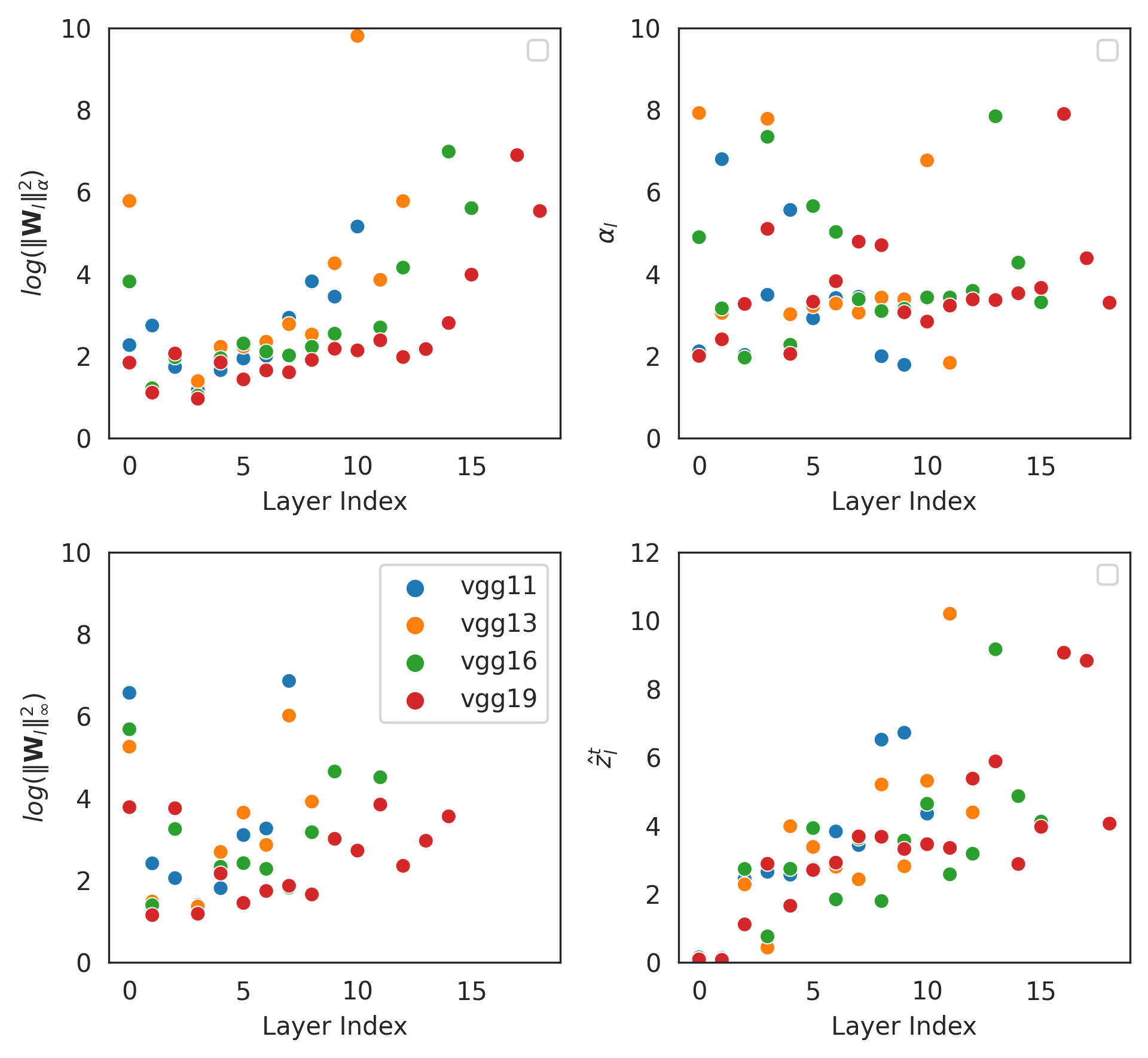}
\captionof{figure}{Comparison between different WeightWatcher (WW) features (left) and \ourmethod (right). Features over layer index for VGGs from pytorchcv of different sizes. \ourmethod shows similar trends to WW, low values at early layers and a sharp increase at the end. }
\vspace{-2mm}
\label{fig:ww_comparison_layers_vgg}
\end{figure}

\begin{figure}[h!]
\centering
\includegraphics[trim=4mm 0mm 1.5mm 1.5mm, width=0.6\linewidth]{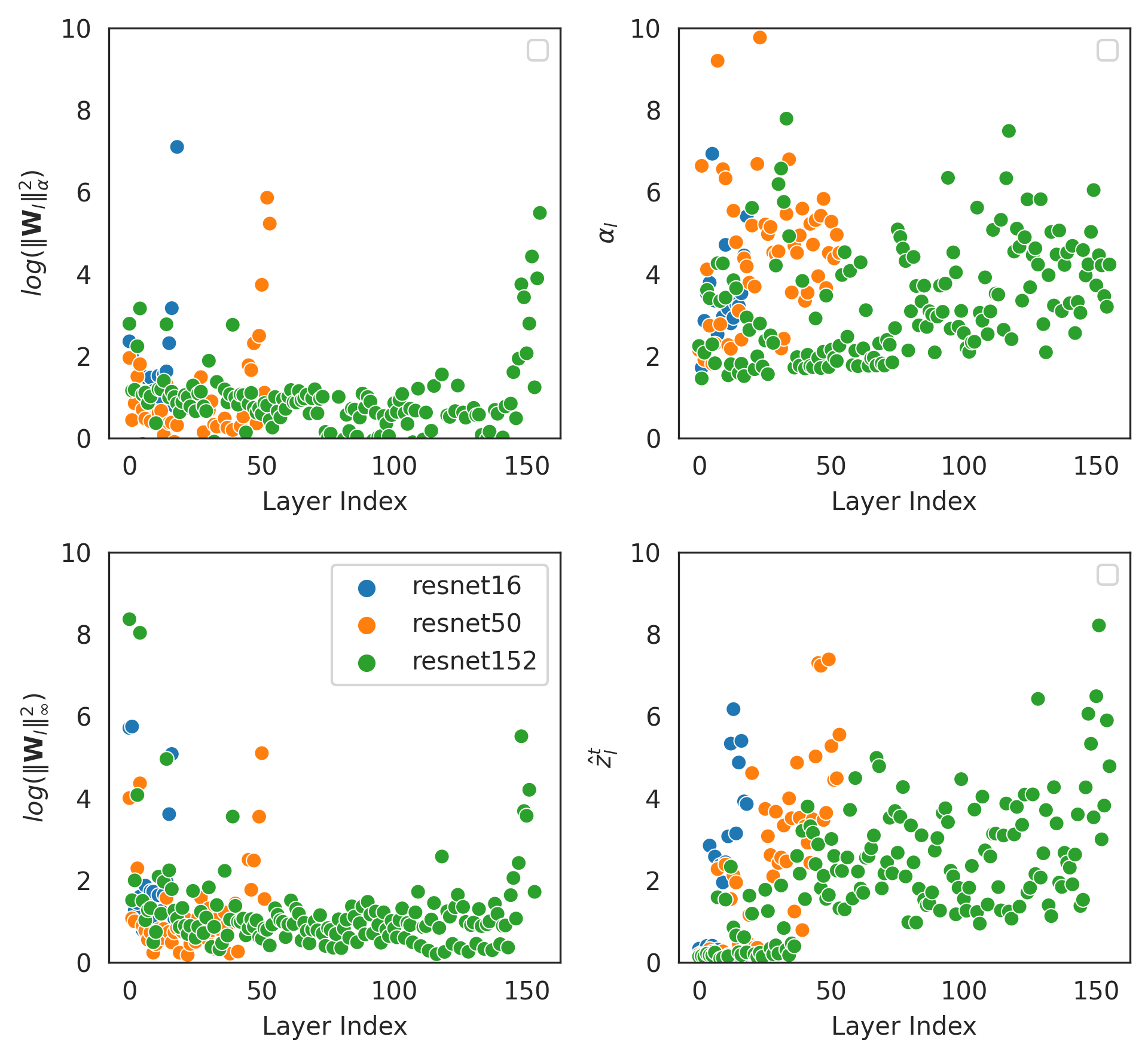}
\captionof{figure}{Comparison between different WeightWatcher (WW) features (left) and \ourmethod (right). Features over layer index for Resnets from pytorchcv of different sizes. \ourmethod shows similar trends to WW, low values at early layers and a sharp increase at the end. }
\vspace{-2mm}
\label{fig:ww_comparison_layers_resnets_2x2}
\end{figure}

Figures \ref{fig:ww_comparison_layers_vgg} and \ref{fig:ww_comparison_layers_resnets_2x2} compare \ourmethod with different WeightWatcher metrics on VGGs from pytorchcv \citep{semeryOsmrImgclsmob2024} and the ResNet-18 zoo from the modelzoo dataset \citep{schurholtModelZoosDataset2022}.
\begin{figure}[h!]
\centering
\includegraphics[trim=4mm 0mm 1.5mm 1.5mm, width=0.6\linewidth]{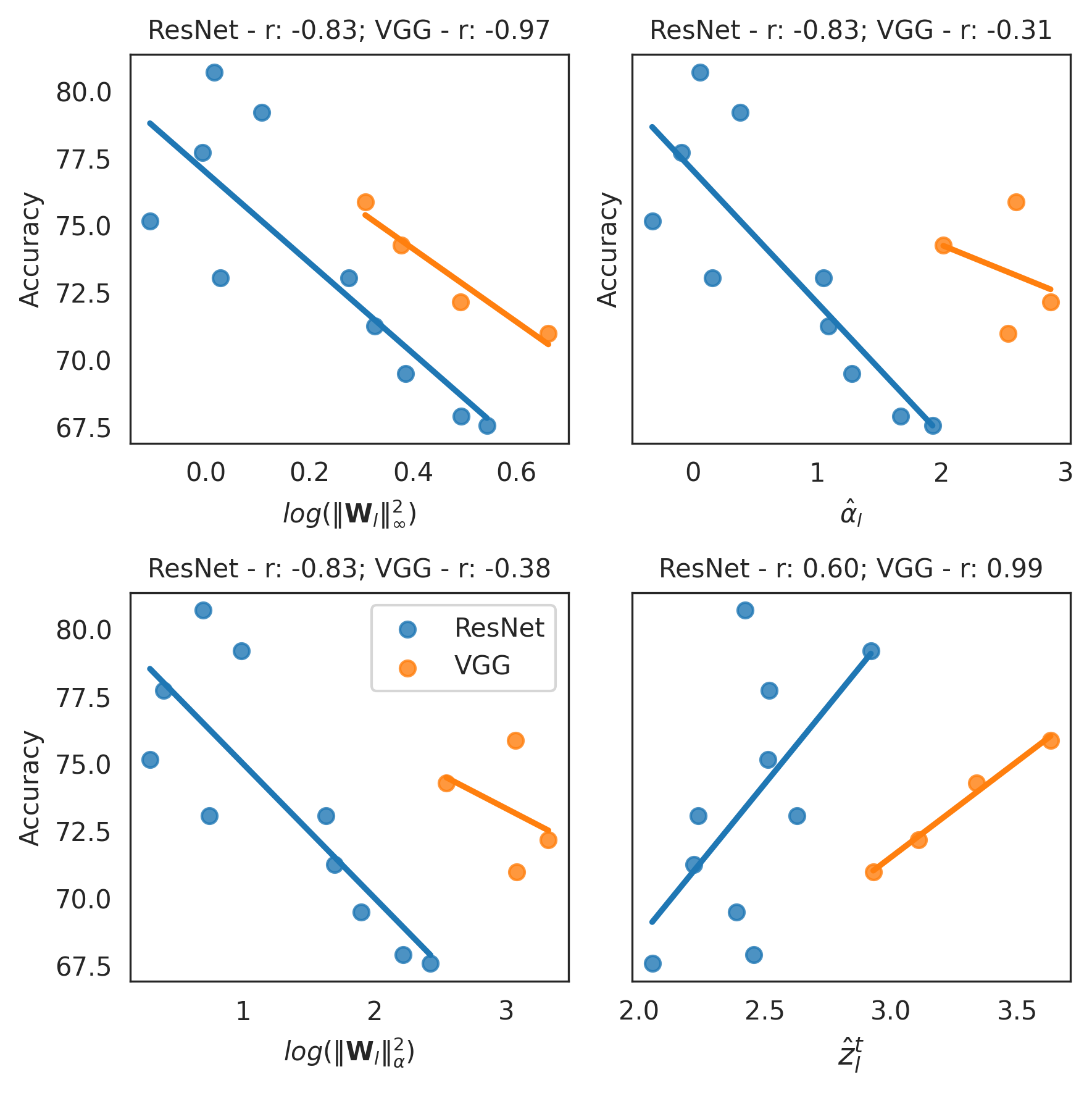}
\captionof{figure}{Comparison between WeightWatcher features (left) and \ourmethod (right). Accuracy over model features for Resnets and VGGs from pytorchcv of different sizes. \ourmethod shows similar trends to WW, low values at early layers and a sharp increase at the end. }
\vspace{-2mm}
\label{fig:ww_comparison_accuracy_2x2}
\end{figure}
\begin{figure}[h!]
\centering
\includegraphics[trim=4mm 0mm 1.5mm 1.5mm, width=0.6\linewidth]{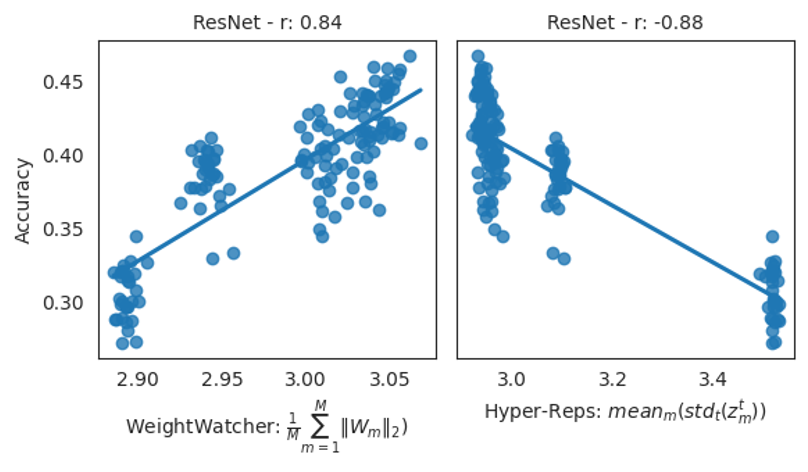}
\captionof{figure}{Comparison between WeightWatcher features (left) and \ourmethod (right). Accuracy over model features for ResNets from the ResNet model zoo. Although \ourmethod is pretrained in a self-supervised fashion, it preserves the linear relation of a globally-aggregated embedding to model accuracy.}
\vspace{-2mm}
\label{fig:ww_comparison_accuracy_our_zoo}    
\end{figure}

\section{Model Property Prediction - Additional Results}
In this section, we provide additional details for Section \ref{sec:property_prediction}. Table \ref{tab:discr_small_zoos_full} shows full results for populations of small CNNs. 
\begin{table}[h!]
\centering
\captionof{table}{
Property prediction on populations of small CNNs. 
}
\label{tab:discr_small_zoos_full}
\scriptsize
\setlength{\tabcolsep}{5pt}
\begin{tabularx}{0.8\linewidth}{lccccccccccccccc}
\toprule
      & \multicolumn{3}{c}{MNIST}                                                      &  & \multicolumn{3}{c}{SVHN}                                                       &  & \multicolumn{3}{c}{CIFAR-10 (CNN)}                                             &  & \multicolumn{3}{c}{STL}                                                        \\ 
      \cmidrule(r){2-4} \cmidrule(lr){6-8} \cmidrule(lr){10-12} \cmidrule(l){14-16} 
      & \multicolumn{1}{c}{W} & \multicolumn{1}{c}{$s(W)$} & \multicolumn{1}{c}{\ourmethod} &  & \multicolumn{1}{c}{W} & \multicolumn{1}{c}{$s(W)$} & \multicolumn{1}{c}{\ourmethod} &  & \multicolumn{1}{c}{W} & \multicolumn{1}{c}{$s(W)$} & \multicolumn{1}{c}{\ourmethod} &  & \multicolumn{1}{c}{W} & \multicolumn{1}{c}{$s(W)$} & \multicolumn{1}{c}{\ourmethod} \\ 
      \cmidrule(r){1-1} \cmidrule(r){2-4} \cmidrule(lr){6-8} \cmidrule(lr){10-12} \cmidrule(l){14-16} 
ACC   & 0.965                 & \textbf{0.987}           & 0.978                       &  & 0.910                 & 0.985                    & \textbf{0.991}              &  & -7.580                & \textbf{0.965}           & 0.885                       &  & -18.818               & \textbf{0.919}           & 0.305                       \\
Epoch & 0.953                 & \textbf{0.974}           & 0.958                       &  & 0.833                 & \textbf{0.953}           & 0.930                       &  & 0.636                 & \textbf{0.923}           & 0.771                       &  & -1.926                & \textbf{0.977}           & 0.344                       \\
Ggap  & 0.246                 & 0.393                    & \textbf{0.402}              &  & 0.479                 & 0.711                    & \textbf{0.760}              &  & 0.324                 & \textbf{0.909}           & 0.772                       &  & -0.617                & \textbf{0.858}           & 0.307                       \\ \bottomrule
\end{tabularx}
\end{table} 

\subsection{Comparison to Previous Work}
\label{app:discr_comparison_previous_work}
Here, we compare \ourmethod with previous work to disseminate the information contained in model embeddings.
The experiment setup in this paper is designed around the ResNets, and therefore it uses sparse epochs for computational efficiency. For consistency, we use the same setup for the CNN zoos as well. The exact numbers are therefore not directly comparable to \citet{schurholtSelfSupervisedRepresentationLearning2021}.   
To provide as much context as possible, we approach the comparison from two angles:
\begin{itemize}
    \item [(1)]\textbf{Direct comparison to the published results:} to contextualize, we use the (deterministic) results of weight statistics $s(W)$ to adjust for the differences in setup. We mark the results for $s(W)$ from \citet{schurholtSelfSupervisedRepresentationLearning2021} as $s(W)_{pp}$ and compare to their $E_{c+}D$ where possible.
    \item [(2)] \textbf{Approximation of the effect of global embeddings:} previous work used global model embeddings, which we approximate by using the full model embedding sequence. We therefore compare \ourmethod + aggregated tokens (as proposed in the submission) to \ourmethod + full model sequence (similar to \citet{schurholtSelfSupervisedRepresentationLearning2021}).
\end{itemize}

The results in the Tables below allow the following conclusions:
\begin{itemize}
    \item [(1)] \textbf{\ourmethod matches the performance of previous work:} The only data available for direct comparison is the MNIST zoo. Here, both in direct comparison and in relation to s(W) cross-relating our results with published numbers, \ourmethod matches published performance of $E_c+D$. On other zoos, $E_c+D$ had similar performance to $s(W)$. We likewise find \ourmethod embeddings to have similar performance to $s(W)$ in our experiments.
    \item [(2)] \textbf{\ourmethod + full sequence improves downstream task performance over the \ourmethod + aggregated sequence:} That indicates that \ourmethod + full sequence contains more information for model prediction. However, both \citet{schurholtSelfSupervisedRepresentationLearning2021} and \ourmethod with full sequence have the disadvantage that they do not scale. With growing models, the representation learner of \citet{schurholtSelfSupervisedRepresentationLearning2021} and the input to the linear probe of \ourmethod + full sequence grow accordingly. \ourmethod + aggregated sequence does lose some information on small models, but scales gracefully to large models and remains competitive.
\end{itemize}

\begin{table}[h!]
\small
\centering
\caption{Property Prediction comparison to previous work on the MNIST-CNN model zoo. We compare our linear probing results from weights $W$, layer-wise quintiles $s(W)$, embeddings from \ourmethod either aggregated into one embedding or using the full sequence to results previously published in \citet{schurholtSelfSupervisedRepresentationLearning2021}. We mark their results for $s(W)$ as $s(W)_{pp}$. Since the experimental setup is not the same, the numbers of $s(W)$ do not match.}
\setlength{\tabcolsep}{8pt}
\begin{tabularx}{0.75\linewidth}{ccccccc}
\toprule
      & \multicolumn{1}{c}{W} & \multicolumn{1}{c}{$s(W)$} & \multicolumn{1}{c}{\ourmethod aggregated} & \ourmethod full sequence & \multicolumn{1}{c}{$s(W)_{pp}$} & \multicolumn{1}{c}{$E_{c+}D$} \\
\cmidrule(r){1-1} \cmidrule(lr){2-2} \cmidrule(lr){3-3} \cmidrule(lr){4-4} \cmidrule(lr){5-5} \cmidrule(lr){6-6} \cmidrule(l){7-7}  
ACC   & 0.965                 & \textbf{0.987}           & 0.978                               & \textbf{0.987}     & 0.977                            & 0.973                              \\
Epoch & 0.953                 & 0.974                    & 0.958                               & \textbf{0.975}     & 0.987                            & 0.989                              \\
Ggap  & 0.246                 & 0.393                    & 0.402                               & \textbf{0.461}     & 0.662                            & 0.667     \\
\bottomrule
\end{tabularx}
\label{tab:discr_comparison_mnist}
\end{table}
\begin{table}[h!]
\small
\centering
\caption{Property Prediction comparison to previous work on the SVHN-CNN model zoo. We compare our linear probing results from weights $W$, layer-wise quintiles $s(W)$, to embeddings from \ourmethod either aggregated into one embedding or using the full sequence. For this zoo, previous results are not available.}
\setlength{\tabcolsep}{8pt}
\begin{tabularx}{0.75\linewidth}{ccccccc}
\toprule
      & \multicolumn{1}{c}{W} & \multicolumn{1}{c}{$s(W)$} & \multicolumn{1}{c}{\ourmethod aggregated} & \ourmethod full sequence & \multicolumn{1}{c}{$s(W)_{pp}$} & \multicolumn{1}{c}{$E_{c+}D$} \\
\cmidrule(r){1-1} \cmidrule(lr){2-2} \cmidrule(lr){3-3} \cmidrule(lr){4-4} \cmidrule(lr){5-5} \cmidrule(lr){6-6} \cmidrule(l){7-7}  
ACC   & 0.910                 & \textbf{0.985}           & 0.991                               & \textbf{0.993}     & \textit{n/a}         & \textit{n/a}         \\
Epoch & 0.833                 & \textbf{0.953}           & 0.930                               & \textbf{0.943}     & \textit{n/a}         & \textit{n/a}         \\
Ggap  & 0.479                 & 0.711                    & 0.760                               & \textbf{0.77}      & \textit{n/a}         & \textit{n/a}        \\
\bottomrule
\end{tabularx}
\label{tab:discr_comparison_svhn}
\end{table}

\begin{table}[h!]
\small
\centering
\caption{Property Prediction comparison to previous work on the CIFAR-CNN(m) model zoo. We compare our linear probing results from weights $W$, layer-wise quintiles $s(W)$, to embeddings from \ourmethod either aggregated into one embedding or using the full sequence. For this zoo, previous results are not available.}
\setlength{\tabcolsep}{8pt}
\begin{tabularx}{0.75\linewidth}{ccccccc}
\toprule
      & \multicolumn{1}{c}{W} & \multicolumn{1}{c}{$s(W)$} & \multicolumn{1}{c}{\ourmethod aggregated} & \ourmethod full sequence & \multicolumn{1}{c}{$s(W)_{pp}$} & \multicolumn{1}{c}{$E_{c+}D$} \\
\cmidrule(r){1-1} \cmidrule(lr){2-2} \cmidrule(lr){3-3} \cmidrule(lr){4-4} \cmidrule(lr){5-5} \cmidrule(lr){6-6} \cmidrule(l){7-7}  
ACC   & -7.580                & \textbf{0.965}           & 0.885                               & \textbf{0.947}     & \textit{n/a}         & \textit{n/a}         \\
Epoch & 0.636                 & \textbf{0.923}           & 0.771                               & \textbf{0.879}     & \textit{n/a}         & \textit{n/a}         \\
Ggap  & 0.324                 & \textbf{0.909}           & 0.772                               & \textbf{0.811}     & \textit{n/a}         & \textit{n/a}        \\
\bottomrule
\end{tabularx}
\label{tab:discr_comparison_cifar}
\end{table}

\section{Model Generation - Additional Results}
This section contains additional results from model sampling experiments, extending Section \ref{sec:generating_models}. 
In Table \ref{tab:generative_cnn_zoos_transfer}, we show results on small CNNs transferring to a new task.
Similarly, Table \ref{tab:generative_resnets_zoos_transfer} shows results on ResNet-18 models for task transfers.
Lastly, Tables \ref{tab:generative_resnets_zoos_fewshot_resnet34}, \ref{tab:generative_resnets_zoos_fewshot_ti_resnet34} and \ref{tab:generative_resnets_zoos_fewshot_resnet34_one_prompt_example} contain additional results for transferring from ResNet-18 CIFAR-100 to ResNet34 and/or Tiny-Imagenet. 
\begin{table}[h!]
\small
\centering
\caption{Model generation on CNN model populations transfer learned on a new task. We compare sampled models at different epochs with models trained from scratch and models fine-tuned from the anchor samples.}
\setlength{\tabcolsep}{8pt}
\begin{tabularx}{0.75\linewidth}{ccccccc}
\toprule
  Method              & \multicolumn{3}{c}{SVHN to MNIST}                            & \multicolumn{3}{c}{CIFAR-10 to STL-10}                       \\
\cmidrule(r){1-1} \cmidrule(lr){2-4} \cmidrule(lr){5-7}
          & Epoch 0            & Epoch 1            & Epoch 25           & Epoch 0            & Epoch 1            & Epoch 25           \\
\cmidrule(r){1-1} \cmidrule(lr){2-2} \cmidrule(lr){3-3} \cmidrule(lr){4-4} \cmidrule(lr){5-5} \cmidrule(lr){6-6} \cmidrule(l){7-7}  
tr.fr.scratch   & $\sim$10 /\%       & 20.6+-1.6          & 83.3+-2.6          & $\sim$10 /\%       & 21.3+-1.6          & 44.0+-1.0          \\
pretrained      & 29.1+-7.2          & 84.1+-2.6          & 94.2+-0.7          & 16.2+-2.3          & 24.8+-0.8          & 49.0+-0.9          \\
$S_{KDE30}$      &  31.8+-5.6          &  86.9+-1.4          &  95.5+-0.4          & n/a          & n/a          & n/a          \\
\cmidrule(r){1-1} \cmidrule(lr){2-2} \cmidrule(lr){3-3} \cmidrule(lr){4-4} \cmidrule(lr){5-5} \cmidrule(lr){6-6} \cmidrule(l){7-7}  
\ourmethod$_{KDE30}$ & \textbf{40.2+-4.8} & 86.7+-1.6          & 94.8+-0.4          & 15.5+-2.3          & 24.9+-1.6          & 49.2+-0.5          \\
\ourmethod$_{SUB}$.  & 37.9+-2.8          & \textbf{88.2+-0.5} & \textbf{95.6+-0.3} & \textbf{17.4+-1.4} & \textbf{25.6+-1.7} & \textbf{49.8+-0.6} \\
\bottomrule
\end{tabularx}
\label{tab:generative_cnn_zoos_transfer}
\end{table}

\begin{table}[]
\scriptsize
\centering
\caption{Model generation on ResNet-18 model populations transferred to a new task. We compare sampled models at different transfer learning epochs with models trained from scratch and models fine-tuned from the same anchor samples.}
\setlength{\tabcolsep}{6pt}
\begin{tabularx}{0.75\linewidth}{clccc}
\toprule
Epoch                  & \multicolumn{1}{c}{Method} & CIFAR-10 to CIFAR-100 & CIFAR-100 to Tiny-Imagenet & Tiny-Imagenet to CIFAR-100 \\
\cmidrule(r){1-1} \cmidrule(lr){2-2} \cmidrule(lr){3-3} \cmidrule(lr){4-4} \cmidrule(l){5-5} 
\multirow{5}{*}{0}     & tr. fr. scratch                  & $\sim$1 /\%                      & $\sim$0.5 /\%                        & $\sim$1 /\%                           \\
                       & Finetuned                  & 1.0+-0.3                     & 0.5+-0.0                             & 1.1+-0.2                              \\
                       & \ourmethod$_{KDE30}$                    & 1.0+-0.3                         & 0.5+-0.1                             & 1.0+-0.2                              \\
                       & \ourmethod$_{SUB}$                   & 1.0+-0.3                         & 0.6+-0.0                             & 1.1+-0.2                              \\
                       & \ourmethod$_{BOOT}$                  & 1.1+-0.2                         & 0.5+-0.0                             & 0.9+-0.2                              \\
\cmidrule(r){1-1} \cmidrule(lr){2-2} \cmidrule(lr){3-3} \cmidrule(lr){4-4} \cmidrule(l){5-5} 
\multirow{5}{*}{1}     & tr. fr. scratch                  & 17.5+-0.7                        & 13.8+-0.8                            & 17.5+-0.7                             \\
                       & Finetuned                  & 27.5+-1.3                     & 25.7+-0.5                            & 51.7+-0.5                             \\
                       & \ourmethod$_{KDE30}$                    & 26.8+-1.4                        & 21.5+-0.9                            & 40.2+-1.0                             \\
                       & \ourmethod$_{SUB}$                   & 26.4+-1.9                        & 21.5+-1.0                            & 40.63+-1.3                            \\
                       & \ourmethod$_{BOOT}$                  & 25.7\_01.9                       & 21.7+-1.0                            & 40.9+-0.8                             \\
\cmidrule(r){1-1} \cmidrule(lr){2-2} \cmidrule(lr){3-3} \cmidrule(lr){4-4} \cmidrule(l){5-5} 
\multirow{5}{*}{5}     & tr. fr. scratch                  & 36.5+-2.0                        & 31.1+-1.6                            & 36.5+-2.0                             \\
                       & Finetuned                  &          45.7+-1.0                        & 36.3+-2.5                            & 52.6+-1.3                             \\
                       & \ourmethod$_{KDE30}$                    & 44.5+-2.0                        & 36.3+-1.2                            & 47.2+-3.3                             \\
                       & \ourmethod$_{SUB}$                   & 45.6+-1.2                        & 35.8+-1.4                            & 49.8+-2.3                             \\
                       & \ourmethod$_{BOOT}$                  & 43.3+-2.4                        & \textbf{37.3+2.0}                    & 50.2+-3.4                             \\
\cmidrule(r){1-1} \cmidrule(lr){2-2} \cmidrule(lr){3-3} \cmidrule(lr){4-4} \cmidrule(l){5-5} 
\multirow{5}{*}{15}    & tr. fr. scratch                  & 53.3+-2.0                        & 38.5+-1.9                            & 53.3+-2.0                             \\
                       & Finetuned                  & 71.9+-0.1                     & 63.4+-0.2                            & \textbf{73.9+-0.3}                    \\
                       & \ourmethod$_{KDE30}$                    & 71.8+-0.3                        & \textbf{63.6+-0.2}                   & 73.4+-0.2                             \\
                       & \ourmethod$_{SUB}$                   & 72.0+-0.2                        & \textbf{63.6+-0.3}                   & 73.5+-0.2                             \\
                       & \ourmethod$_{BOOT}$                  & 71.9+-0.3                        & 63.4+-0.1                            & 73.7+-0.3                             \\
\cmidrule(r){1-1} \cmidrule(lr){2-2} \cmidrule(lr){3-3} \cmidrule(lr){4-4} \cmidrule(l){5-5} 
25 & tr. fr. scratch                  & 56.5+-2.0                        & 43.3+-1.9                            & 56.5+-2.0                             \\
50 & tr. fr. scratch                  & 70.7+-0.4                        & 57.3+-0.6                            & 70.7+-0.4                             \\
60 & tr. fr. scratch                  & 74.2+-0.3                        & 63.9+-0.5                            & 74.2+-0.3                         \\
\bottomrule
\end{tabularx}
\label{tab:generative_resnets_zoos_transfer}
\end{table}

\begin{table}[h]
\vspace{-4mm}
\centering
\captionof{table}{
Few-shot model generation for a new task: Sampling ResNet-34 models for CIFAR-100. \ourmethod was pretrained on CIFAR-100 ResNet-18s, 5 samples are drawn using subsampling. To get prompt examples, we train 3 ResNet-34 models on CIFAR-100 for 2 epochs to a mean accuracy of 26 \%. }
\label{tab:generative_resnets_zoos_fewshot_resnet34}
\small
\setlength{\tabcolsep}{7pt}
\begin{center}
\begin{tabularx}{0.4\linewidth}{clcc}
\toprule
\multicolumn{4}{c}{CIFAR100   ResNet-18 to ResNet-34}                 \\
\midrule
  Ep.   & \multicolumn{1}{c}{Method}  & 5 Epochs           & 15 Epochs          \\
\cmidrule(r){1-1}  \cmidrule(l){2-2} \cmidrule(l){3-4} 
0  & tr. fr. Scratch           & 1.0$\pm$0.1           & 1.0$\pm$0.1           \\
         & \ourmethod & \textbf{1.6$\pm$0.3}  & \textbf{1.6$\pm$0.3}  \\
\cmidrule(r){1-1}  \cmidrule(l){2-2} \cmidrule(l){3-4} 
 1  & tr. fr. Scratch           & 12.4$\pm$1.0          & 12.9$\pm$0.8          \\
         & \ourmethod & \textbf{16.8$\pm$0.7} & \textbf{23.1$\pm$0.3} \\
\cmidrule(r){1-1}  \cmidrule(l){2-2} \cmidrule(l){3-4} 
5  & tr. fr. Scratch           & 49.5$\pm$0.6          & 36.2$\pm$1.7          \\
         & \ourmethod & \textbf{51.9$\pm$0.6} & \textbf{37.8$\pm$1.4} \\
\cmidrule(r){1-1}  \cmidrule(l){2-2} \cmidrule(l){3-4} 
15 & tr. fr. scratch      &      & 68.8$\pm$0.4          \\
         & \ourmethod &                    & \textbf{69.3$\pm$0.3} \\
\cmidrule(r){1-1}  \cmidrule(l){2-2} \cmidrule(l){3-4} 
 & \ourmethod Ens.                & 53.5               & 71.3         \\
\bottomrule
\end{tabularx}
\end{center}
\vspace{-4mm}
\end{table} 
\begin{table}[h]
\vspace{-4mm}
\small
\captionof{table}{
Few-shot model generation for a new task and architecture: \ourmethod trained on CIFAR-100 ResNet-18s used to generate ResNet-34s for Tiny-Imagenet. 5 samples are drawn using subsampling. To get prompt examples, we train 3 ResNet-34 models on Tiny-Imagenet for 2 epochs to a mean accuracy of 28.5 \%. 
}
\label{tab:generative_resnets_zoos_fewshot_ti_resnet34}
\centering
\begin{center}
\setlength{\tabcolsep}{7pt}
\begin{tabularx}{0.45\linewidth}{clcc}
\multicolumn{4}{c}{ResNet-18   CIFAR100 to ResNet-34 Tiny-Imagenet}                           \\
\cmidrule(r){1-1}  \cmidrule(l){2-2} \cmidrule(l){3-4} 
Ep.               & Method                    & 5 epochs           & 15 epochs          \\
\cmidrule(r){1-1}  \cmidrule(l){2-2} \cmidrule(l){3-4} 
\multirow{2}{*}{0}  & tr. fr. Scratch           & 0.5$\pm$0.0           & 0.5$\pm$0.0           \\
                    & \ourmethod & 0.5$\pm$0.1           & 0.6$\pm$0.2           \\
\cmidrule(r){1-1}  \cmidrule(l){2-2} \cmidrule(l){3-4} 
\multirow{2}{*}{1}  & tr. fr. Scratch           & 10.5$\pm$1.4          & 11.9$\pm$1.9          \\
                    & \ourmethod & \textbf{13.3$\pm$0.5} & \textbf{18.5$\pm$0.7} \\
\cmidrule(r){1-1}  \cmidrule(l){2-2} \cmidrule(l){3-4} 
\multirow{2}{*}{5}  & tr. fr. Scratch           & 47.2$\pm$0.7          & 31.1$\pm$1.7          \\
                    & \ourmethod & \textbf{50.6$\pm$0.3} & \textbf{31.6$\pm$0.6} \\
\cmidrule(r){1-1}  \cmidrule(l){2-2} \cmidrule(l){3-4} 
\multirow{2}{*}{15} & \multicolumn{2}{l}{tr. fr. Scratch}            & 61.9$\pm$0.3          \\
                    & \ourmethod &                    & \textbf{62.7$\pm$0.3} \\
\cmidrule(r){1-1}  \cmidrule(l){2-2} \cmidrule(l){3-4} 
            & \ourmethod Ens.                           & 52                 & 65.1              
\\
\bottomrule
\end{tabularx}
\end{center}
\vspace{-4mm}
\end{table}

\subsection{Diversity of sampled models}
\label{sec:sampling_diversity}
An interesting question is whether sampling \ourmethod generates versions of the same model. To test that,  we evaluate the diversity of samples generated with only a few few-shot examples by combining the models to ensembles. 
The improvements of the ensembles over the individual models demonstrate their diversity.
This indicates that given very few, early-stage prompt examples, sampling hyper-representations improves learning speed and performance in otherwise equal settings. 
Additionally, we conducted experiments with varying numbers of prompt examples, revealing that increasing the number of prompt examples enhances both performance and diversity. 
Nonetheless, even a single prompt example trained for just 2 epochs contains sufficient information to generate model samples that surpass those derived from random initialization; see Table \ref{tab:generative_resnets_zoos_fewshot_resnet34_one_prompt_example}.

\begin{table}[h]
\centering
\captionof{table}{
Sampling ResNet-34 models for CIFAR-100. \ourmethod was pretrained on CIFAR-100 ResNet-18s, 5 samples are drawn using subsampling. To get prompt examples, we train a single ResNet-34 model on CIFAR-100 for 2 epochs to an accuracy of 26 \%. 
}
\label{tab:generative_resnets_zoos_fewshot_resnet34_one_prompt_example}
\small
\begin{center}
\setlength{\tabcolsep}{7pt}
\begin{tabularx}{0.5\linewidth}{clcc}
\toprule
\multicolumn{4}{c}{CIFAR100   ResNet-18 to ResNet-34}                 \\
\cmidrule(r){1-1}  \cmidrule(l){2-2} \cmidrule(l){3-4} 
  Epoch   & \multicolumn{1}{c}{Method}  & 5 Epochs           & 15 Epochs          \\
\cmidrule(r){1-1}  \cmidrule(l){2-2} \cmidrule(l){3-4} 
0  & tr. fr. Scratch            & 1.0+-0.1           & 1.0+-0.1           \\
         & \ourmethod           & \textbf{1.5+-0.2}  & \textbf{1.6+-0.1}  \\
\cmidrule(r){1-1}  \cmidrule(l){2-2} \cmidrule(l){3-4} 
 1  & tr. fr. Scratch           & 12.4+-1.0          & 12.9+-0.8          \\
         & \ourmethod           & \textbf{16.9+-0.7} & \textbf{19.4+-0.2} \\
\cmidrule(r){1-1}  \cmidrule(l){2-2} \cmidrule(l){3-4} 
5  & tr. fr. Scratch            & 49.5+-0.6          & 36.2+-1.7          \\
         & \ourmethod           & \textbf{51.5+-0.3} & \textbf{38.6+-1.6} \\
\cmidrule(r){1-1}  \cmidrule(l){2-2} \cmidrule(l){3-4} 
15 & tr. fr. scratch            &      &             68.8+-0.4          \\
         & \ourmethod &                              & \textbf{69.1+-0.1} \\
\cmidrule(r){1-1}  \cmidrule(l){2-2} \cmidrule(l){3-4} 
Ensemble & \ourmethod                & 51.8          & 70.2              
\\
\bottomrule
\end{tabularx}
\end{center}
\end{table} 


\end{document}